\documentclass[lettersize,journal]{IEEEtran}
\usepackage{amsmath,amsfonts}
\usepackage{algorithmic}
\usepackage{algorithm}
\usepackage{array}
\usepackage[caption=false,font=normalsize,labelfont=sf,textfont=sf]{subfig}
\usepackage{textcomp}
\usepackage{stfloats}
\usepackage{url}
\usepackage{verbatim}
\usepackage{graphicx}
\usepackage{booktabs}
\usepackage{multirow}
\usepackage{cite}
\begin{document}
\title{Fresco++: Frequency-Guided and Canonical-Consistent Optimization for Fine-Grained Head Avatar Modeling}

\title{Fresco++: Frequency-Guided and Canonical-Consistent Optimization for Fine-Grained Head Avatar Modeling}

\author{
Shikun Zhang,
Yong Li,
Yiqun Wang,
and Cunjian Chen$^{*}$, Senior Member, IEEE
\thanks{Shikun Zhang, and Cunjian Chen are with the Department of Data Science and AI, Monash University, Australia
.}%
\thanks{Yong Li and Yiqun Wang are with the College of Computer Science, Chongqing University, China
.}%
\thanks{$^{*}$Corresponding author: Cunjian Chen.}%
}









\maketitle

\begin{abstract}
We propose Fresco++, a unified optimization framework for fine-grained and view-consistent head avatar reconstruction. Head avatar optimization is typically driven by per-view image supervision, which can lead to premature fitting of unstable high-frequency details and inconsistent local appearance across viewpoints. Fresco++ addresses these challenges by regulating both the progression of visual detail and the formation of cross-view supervision during optimization. For frequency-aware optimization, a progressive curriculum first stabilizes low-frequency structures and then introduces high-frequency constraints to recover fine facial and hair details without amplifying spurious responses at early stages. For cross-view optimization, we introduce Canonical Group Consensus, which associates local observations through shared canonical surface regions and establishes correspondence across different viewpoints. Geometric and visibility-aware screening removes unreliable observations, while the remaining multi-view evidence is aggregated in feature space to form a consensus target for supervising the current rendering. This design enforces local consistency without relying on a specific image-space parameterization and avoids additional rendering of the auxiliary view. Together, the frequency curriculum and canonical consensus provide stable optimization from coarse structures to fine details while maintaining coherent appearance across viewpoints. Extensive experiments on NeRSemble demonstrate improved reconstruction quality and cross-view consistency, while evaluations across diverse avatar representations further confirm the generality and transferability of Fresco++.
\end{abstract}


\section{Introduction}
\label{sec:intro}

Generating animatable human head avatars has long been a central research topic in computer vision and graphics. 
High-quality avatars are expected to faithfully reproduce identity-specific appearance and fine-grained facial dynamics while remaining controllable under varying expressions, head poses, and viewpoints. 
Such capability is essential for applications including gaming, film production, immersive communication, and AR/VR~\cite{waggoner2009my,he2020collabovr,lombardi2018deep,orts2016holoportation}. 
Despite substantial progress in neural rendering, simultaneously achieving fine visual fidelity, stable optimization, and coherent appearance across viewpoints remains challenging.

Recent head avatar methods have explored increasingly diverse representations, ranging from implicit neural fields~\cite{mildenhall2021nerf,yariv2021volume,wang2021neus,barron2021mip} to explicit meshes, Gaussian primitives, and hybrid representations~\cite{qian2024gaussianavatars,xiang2024flashavatar,wang2025mega,zhang2025hravatar,feng2025gpavatar,sun2025svghead}. 
In particular, recent Gaussian-based approaches have substantially improved rendering efficiency and representation flexibility through learnable deformation, high-dimensional Gaussian embeddings, and hybrid surface--volumetric modeling~\cite{zhang2025hravatar,feng2025gpavatar,sun2025svghead}. 
However, high-fidelity reconstruction still requires multiple geometric, appearance, deformation, opacity, and view-dependent components to be jointly optimized under image-space supervision. 
Such highly coupled optimization remains susceptible to unstable local solutions: fine-scale responses may be fitted before reliable coarse structures have formed, while local appearance or structure may drift when the same facial region is observed from different viewpoints~\cite{liao2025hhavatar,xu2025gphm,wang20253d,qin2024high}.

Existing approaches have addressed these challenges from several complementary directions. 
Frequency-aware methods regulate the spectral content exposed during training to reduce premature fitting of high-frequency signals~\cite{lin2021barf,yang2023freenerf,zhang2024fregs}, while recent reconstruction and avatar methods have further adopted progressive or coarse-to-fine learning strategies to improve hierarchical detail recovery~\cite{ren2026omgavatar}. 
For cross-view coherence, recent head-avatar methods exploit multi-view generative priors or explicit consistency constraints to improve novel-view stability~\cite{tang2025gaf,chharia2026mvchead}. 
Meanwhile, structural correspondence is commonly established through representation-specific mechanisms such as template-bound coordinates, UV parameterizations, or surface-local Gaussian associations~\cite{sun2025svghead,prinzler2026match,song2026progressiveavatars}. 
These strategies have demonstrated the importance of optimization scheduling, multi-view supervision, and structural correspondence, but they are typically instantiated as separate mechanisms tied to the formulation of the underlying representation. 
Consequently, a central challenge remains: how to regulate the progression of fine-detail learning while establishing reliable cross-view supervision through a common correspondence unit that can be transferred across different avatar representations.

To address this challenge, we propose Fresco++, a unified optimization framework for fine-grained and view-consistent head avatar reconstruction. 
We first regulate the evolution of image details through a spatially localized frequency curriculum. 
Rather than exposing the model to all frequency components throughout training, Fresco++ emphasizes low-frequency structures during early optimization, suppressing unstable fine-scale responses before the overall facial appearance and structure become reliable. 
As optimization progresses, high-frequency supervision is progressively emphasized to recover expression-sensitive facial details, boundaries, and hair structures. 
This coarse-to-fine progression provides an ordered optimization trajectory in which structural stabilization precedes detail refinement.

Beyond regulating when different levels of detail should be learned, Fresco++ further addresses where cross-view supervision should be established and how reliable evidence should be formed. 
To this end, we introduce Canonical Group Consensus (CGC). 
CGC uses local structural identities defined in a shared canonical domain to associate corresponding observations across synchronized viewpoints, providing a common correspondence unit for cross-view supervision.
The corresponding observations of each canonical group are then collected from multiple views, while unreliable observations caused by unfavorable geometry or visibility are excluded from supervision. 
Instead of directly enforcing pairwise agreement between different view predictions, the remaining multi-view evidence is consolidated into a feature-level consensus to guide the current reconstruction. 
In this way, CGC integrates correspondence construction, reliability assessment, and cross-view supervision within a unified canonical formulation, without binding the consistency objective to a particular image-space parameterization.

The frequency curriculum and Canonical Group Consensus address two complementary aspects of the reconstruction process. 
The former regulates the temporal progression from stable coarse structures toward fine details, while the latter provides structured and reliable spatial supervision across viewpoints. 
Together, they form a representation-transferable optimization strategy that improves reconstruction fidelity and cross-view coherence without requiring a representation-specific consistency formulation.

A preliminary version of this work, Fresco~\cite{zhang2026fresco}, appeared at CVPR 2026 (Highlight).
The present journal version substantially extends the original formulation. 
Most importantly, we generalize cross-view consistency from UV-space texel alignment to Canonical Group Consensus, introducing canonical local-group correspondence, reliability-aware multi-view supervision, and feature-level consensus construction. 
We further provide a substantially expanded evaluation across different avatar representations, together with additional analyses of the optimization behavior and design choices of the proposed framework. 
The resulting formulation is presented as Fresco++ in this work.

Our main contributions are summarized as follows:
\begin{itemize}
   \item Building upon Fresco, we propose Fresco++, a unified and representation-transferable optimization framework for fine-grained and view-consistent head avatar reconstruction.
   It coordinates coarse-to-fine frequency progression with reliable cross-view supervision, while removing Fresco's dependence on a specific representation, thereby enabling its application across different avatar representations.

    \item We introduce Canonical Group Consensus (CGC), which establishes local structural correspondence through a shared canonical reference, identifies reliable multi-view observations, and consolidates them into feature-level consensus supervision without relying on a specific image-space parameterization.

    \item Extensive experiments demonstrate improved reconstruction fidelity and cross-view coherence. Evaluations across different avatar representations, together with detailed optimization and ablation analyses, further validate the effectiveness and transferability of Fresco++.
\end{itemize}

\section{Related Work}
\label{sec:related works}

\subsection{Representations for Animatable Head Avatars}

Animatable head avatars have been modeled using implicit neural fields, explicit meshes, Gaussian primitives, and hybrid representations.
Implicit approaches encode geometry and appearance in continuous spaces and achieve high-quality rendering~\cite{gafni2021dynamic,hong2022headnerf,zheng2022avatar}, whereas mesh-based methods exploit explicit surface topology to provide semantic correspondence and controllable deformation~\cite{li2017learning,grassal2022neural,lu2025lsf,zhang2023sdrnet}.
The emergence of 3D Gaussian Splatting provides a flexible explicit alternative with efficient differentiable rendering.
PointAvatar~\cite{zheng2023pointavatar}, GaussianAvatars~\cite{qian2024gaussianavatars}, Gaussian Head Avatar~\cite{wang2025gaussianhead}, HeadGaS~\cite{dhamo2024headgas}, and EAvatar~\cite{zhang2025eavatar} associate Gaussian primitives with facial deformation models in different ways.
Recent methods further improve representation flexibility through learned deformation and more expressive Gaussian parameterizations, as exemplified by HRAvatar~\cite{zhang2025hravatar} and GPAvatar~\cite{feng2025gpavatar}.

Hybrid representations further combine explicit geometric structures with flexible appearance primitives.
MeGA~\cite{wang2025mega} and HERA~\cite{cai2024hybrid} employ mesh--Gaussian formulations to model different head components, while PhysHead~\cite{kabadayi2026physhead} extends this direction with a layered representation combining parametric facial geometry, strand-based hair, and Gaussian appearance primitives.
These developments lead to increasingly diverse parameterizations of geometry, appearance, and deformation, making optimization and supervision mechanisms that remain effective across different avatar representations increasingly important.

\subsection{Optimization for Stable and Fine-Grained Reconstruction}

Reconstruction quality depends not only on representation capacity but also on how different levels of information are introduced during optimization.
Frequency-aware strategies have shown that progressively exposing high-frequency signals can improve optimization stability.
BARF~\cite{lin2021barf} and Nerfies~\cite{park2021nerfies} employ frequency annealing for coarse-to-fine optimization, while FreeNeRF~\cite{yang2023freenerf} regulates high-frequency components under sparse-view supervision.
Other approaches, including G$^2$FR~\cite{xie2024g}, SpectralNeRF~\cite{li2024spectralnerf}, and FreGS~\cite{zhang2024fregs}, explicitly exploit spectral information to regularize neural or Gaussian-based reconstruction.
In particular, FreGS progressively expands effective frequency bands during Gaussian Splatting optimization, demonstrating the relevance of frequency scheduling to explicit primitive-based representations.

Beyond explicit spectral regulation, recent Gaussian-avatar methods also adopt progressive or coarse-to-fine optimization.
TetGS organizes avatar reconstruction into successive optimization stages and employs coarse-to-fine appearance refinement~\cite{liu2025tetgs}, while OMG-Avatar uses a coarse-to-fine learning paradigm to recover hierarchical details in a multi-level-of-detail Gaussian representation~\cite{ren2026omgavatar}.
Nevertheless, these strategies primarily regulate \emph{when} and \emph{how} different levels of detail are learned; they do not explicitly determine whether observations from different viewpoints correspond to the same local structure or provide sufficiently reliable cross-view supervision.

\subsection{Cross-View Consistency and Structural Correspondence}

Cross-view consistency has become increasingly important in head avatar reconstruction and generation, particularly when observations are sparse or unevenly distributed across viewpoints.
Recent methods address this issue through additional multi-view priors or explicit consistency constraints.
GAF~\cite{tang2025gaf} distills predictions from a multi-view head diffusion model as pseudo-ground-truth observations to regularize monocular Gaussian-avatar reconstruction and improve novel-view consistency.
MVCHead~\cite{chharia2026mvchead}, under a different generative setting, introduces a multi-view critic to encourage consistency among self-rendered views without requiring real multi-view image pairs.
These approaches demonstrate that incorporating cross-view information beyond independent per-view supervision can substantially improve view consistency, although the corresponding mechanisms are often coupled with particular generative priors or training objectives.

A complementary line of research focuses on establishing stable structural correspondence for explicit avatar representations.
SVG-Head~\cite{sun2025svghead} uses mesh-aware Gaussian UV mapping to associate surface Gaussians with a FLAME-based structure.
MATCH~\cite{prinzler2026match} explicitly models topological correspondence among Gaussian splats for multi-view head-avatar reconstruction and editing.
ProgressiveAvatars~\cite{song2026progressiveavatars} instead defines Gaussian primitives in face-local coordinates on a template mesh to preserve local associations under deformation.
Together, these approaches demonstrate several effective correspondence mechanisms, including UV parameterization, topological registration, and surface-local coordinates.

Despite these advances, structural correspondence and cross-view supervision are often developed for different purposes.
Structural associations mainly support representation construction, deformation, registration, or editing, whereas cross-view consistency is frequently enforced through task-specific priors or dedicated consistency objectives.
A more general formulation that uses stable structural correspondence to organize reliable multi-view supervision therefore remains underexplored.

\begin{figure*}[t]
  \centering
  
  \includegraphics[width=0.95\textwidth]{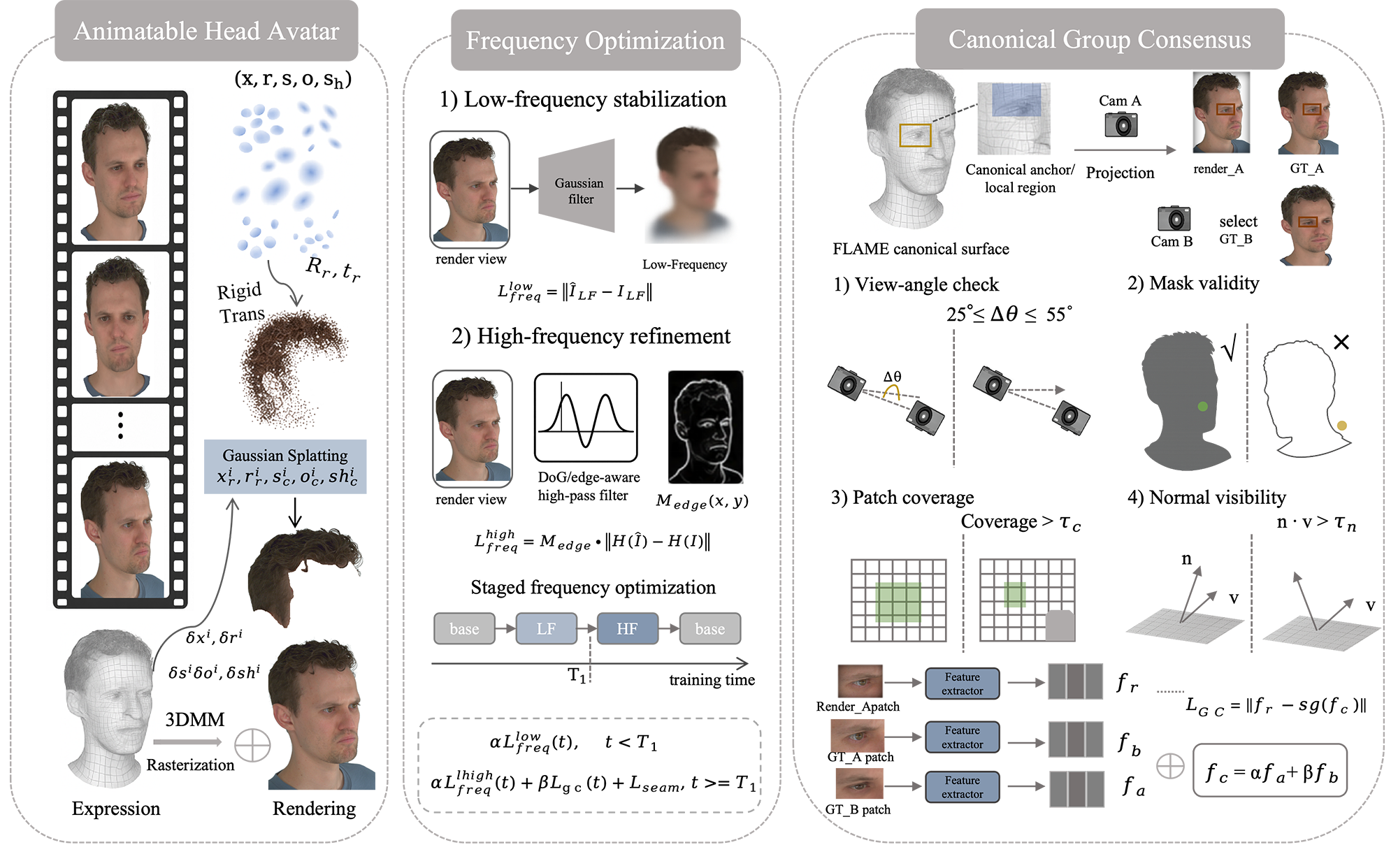}

   \caption{Overview of the proposed Fresco++.
Multi-view images are used to estimate the facial 3DMM and refine the Gaussian hair parameters, while the facial mesh and Gaussian field are jointly rendered to produce the final head avatar.
In the \emph{frequency optimization} branch, low-frequency supervision stabilizes coarse appearance and structure, followed by edge-aware high-frequency refinement under a staged low-to-high frequency schedule.
In the \emph{Canonical Group Consensus (CGC)} branch, local regions are associated through shared canonical FLAME anchors and projected to multiple views.
Reliable observations are selected through view-angle, mask-validity, patch-coverage, and normal-visibility checks, after which multi-view GT features are aggregated into a canonical consensus target to supervise the rendered patch.
The two branches are jointly integrated into the overall optimization to improve fine-grained reconstruction and cross-view consistency.}
   \label{fig:network}
\end{figure*}

\section{Method}
\label{sec:method}

We propose Fresco++, a unified optimization framework for fine-grained and view-consistent head avatar reconstruction.
Rather than redesigning the underlying avatar representation, Fresco++ focuses on the optimization process from two complementary aspects: the progression of visual detail and the construction of reliable cross-view supervision.
Specifically, a spatially localized frequency curriculum guides the reconstruction from coarse structures toward fine details, while Canonical Group Consensus (CGC) establishes structured correspondence in a shared canonical domain and consolidates reliable multi-view observations into consistent supervision.
These two components jointly regulate the temporal and cross-view behavior of the reconstruction while preserving the original deformation and rendering pipeline of the underlying avatar model.
An overview of the complete framework is illustrated in Fig.~\ref{fig:network}.

\subsection{Preliminaries}
\label{sec:preliminary}

Fresco++ formulates fine-grained head avatar reconstruction as a joint optimization problem over appearance detail progression and cross-view structural consistency.
In our primary implementation, the head is represented using a hybrid mesh--Gaussian formulation, where the facial region is modeled by a parametric FLAME surface and the hair and peripheral regions are represented by 3D Gaussian primitives.
This formulation combines explicit facial structure with the flexible modeling capability of Gaussian primitives for regions with complex geometry and appearance.

Let
\begin{equation}
\mathcal{M}_t=
\mathcal{M}(\boldsymbol{\beta},\boldsymbol{\psi}_t,\boldsymbol{\phi}_t)
\end{equation}
denote the facial mesh at frame $t$, where $\boldsymbol{\beta}$, $\boldsymbol{\psi}_t$, and $\boldsymbol{\phi}_t$ represent identity shape, expression, and pose, respectively.
Its canonical configuration is denoted by $\mathcal{M}_c$, and the underlying deformation model maps $\mathcal{M}_c$ to the posed surface $\mathcal{M}_t$ according to the current expression and pose.

The Gaussian component is represented as
\begin{equation}
\mathcal{G}
=
\left\{
(\boldsymbol{\mu}_i,\boldsymbol{\Sigma}_i,\mathbf{c}_i,\alpha_i)
\right\}_{i=1}^{N},
\end{equation}
where $\boldsymbol{\mu}_i\in\mathbb{R}^{3}$ denotes the center of the $i$-th Gaussian,
$\boldsymbol{\Sigma}_i$ its anisotropic covariance,
$\mathbf{c}_i$ its appearance attributes, and
$\alpha_i$ its opacity.
The Gaussian primitives are transformed according to the deformation mechanism of the underlying avatar model so that their spatial configuration follows the current head state.

Given camera parameters $\mathbf{\Pi}_{t}^{v}$ for view $v$, the facial surface is rasterized and the Gaussian primitives are splatted into the image plane.
We denote the resulting rendered image as
\begin{equation}
\hat{\mathbf{I}}_{t}^{v}
=
\mathcal{R}
\left(
\mathcal{A}_{t},
\mathbf{\Pi}_{t}^{v}
\right),
\end{equation}
where $\mathcal{A}_{t}$ denotes the complete posed avatar at frame $t$ and
$\mathcal{R}(\cdot)$ represents the differentiable rendering process.
The corresponding captured image is denoted by $\mathbf{I}_{t}^{v}$ and provides the standard image-domain supervision.

For synchronized multi-view capture, the observations
$\{\mathbf{I}_{t}^{v}\}_{v=1}^{V}$
share the same underlying facial state $\mathcal{A}_{t}$ and differ only in their camera viewpoints.
Through the deformation mapping between the canonical surface $\mathcal{M}_{c}$ and the posed surface $\mathcal{M}_{t}$, local structures observed from different views can be related to a common canonical reference.
This shared relationship between the canonical structure and synchronized multi-view observations provides the basis for establishing cross-view correspondence in the subsequent optimization, while the rendered observations are directly used for image-domain supervision.

\subsection{Spatially Localized Frequency Curriculum}
\label{sec:frequency}

Fine-grained head reconstruction requires the model to capture both stable large-scale structures and subtle local details.
When all frequency components are optimized without distinction, rapidly varying image signals may dominate before coarse appearance and structure are sufficiently established, leading to unstable detail fitting.
To better organize the evolution of visual information during reconstruction, we introduce a spatially localized frequency curriculum that separates low-frequency stabilization from high-frequency refinement.
Unlike global spectral regulation, our formulation operates directly in the image domain and preserves the spatial location of frequency responses, which is particularly important for localized facial structures and hair details.

\noindent\textbf{Low-Frequency Stabilization.}
Given a rendered image $\hat{\mathbf{I}}$ and its corresponding ground-truth image $\mathbf{I}$, we extract their low-frequency components using Gaussian filtering:
\begin{equation}
\hat{\mathbf{I}}_{\mathrm{LF}}
=
\mathcal{G}_{\sigma}(\hat{\mathbf{I}}),
\qquad
\mathbf{I}_{\mathrm{LF}}
=
\mathcal{G}_{\sigma}(\mathbf{I}),
\label{eq:low_frequency}
\end{equation}
where $\mathcal{G}_{\sigma}(\cdot)$ denotes Gaussian filtering with standard deviation $\sigma$.
The corresponding low-frequency reconstruction objective is
\begin{equation}
\mathcal{L}_{\mathrm{LF}}
=
\left\|
\hat{\mathbf{I}}_{\mathrm{LF}}
-
\mathbf{I}_{\mathrm{LF}}
\right\|_{1}.
\label{eq:lf_loss}
\end{equation}
By attenuating rapidly varying responses, $\mathcal{L}_{\mathrm{LF}}$ emphasizes spatially coherent appearance and coarse structural information, providing a more stable basis before fine-scale details are strongly optimized.

\noindent\textbf{High-Frequency Refinement.}
As the reconstruction progresses, fine structures such as facial boundaries, wrinkles, and hair details require increasingly stronger high-frequency supervision.
We extract localized band-pass responses using a Difference-of-Gaussians (DoG) operator:
\begin{equation}
\mathcal{H}(\mathbf{I})
=
\mathcal{G}_{\sigma_1}(\mathbf{I})
-
\mathcal{G}_{\sigma_2}(\mathbf{I}),
\qquad
\sigma_1 < \sigma_2,
\label{eq:dog}
\end{equation}
where $\sigma_1$ and $\sigma_2$ determine the frequency range emphasized by the filter.
Since the DoG response remains spatially localized, high-frequency supervision can be associated with specific facial and hair regions rather than only global spectral statistics.

We further emphasize structurally informative regions using an edge-aware weighting map $\mathbf{M}_{\mathrm{edge}}$, yielding
\begin{equation}
\mathcal{L}_{\mathrm{HF}}^{\mathrm{edge}}
=
\left\|
\mathbf{M}_{\mathrm{edge}}
\odot
\left(
\mathcal{H}(\hat{\mathbf{I}})
-
\mathcal{H}(\mathbf{I})
\right)
\right\|_{1},
\label{eq:hf_edge}
\end{equation}
where $\odot$ denotes element-wise multiplication.
To preserve local gradient structure, we additionally employ a gradient-difference objective:
\begin{equation}
\mathcal{L}_{\mathrm{GDL}}
=
\sum_{p}
\left|
\left\|\nabla \hat{\mathbf{I}}_{p}\right\|
-
\left\|\nabla \mathbf{I}_{p}\right\|
\right|,
\label{eq:gdl}
\end{equation}
where $p$ indexes image pixels.
The complete high-frequency objective is defined as
\begin{equation}
\mathcal{L}_{\mathrm{HF}}
=
\lambda_{\mathrm{edge}}
\mathcal{L}_{\mathrm{HF}}^{\mathrm{edge}}
+
\lambda_{\mathrm{gdl}}
\mathcal{L}_{\mathrm{GDL}},
\label{eq:hf_loss}
\end{equation}
where $\lambda_{\mathrm{edge}}$ and $\lambda_{\mathrm{gdl}}$ balance localized frequency reconstruction and gradient preservation.

\noindent\textbf{Stage-Wise Frequency Scheduling.}
Rather than applying low- and high-frequency objectives simultaneously throughout optimization, we organize them through a stage-wise curriculum.
Each stage maintains a fixed frequency emphasis, allowing the reconstruction to first consolidate low-frequency structures and subsequently shift toward high-frequency refinement.
Formally, the frequency-specific objective at iteration $t$ is written as
\begin{equation}
\mathcal{L}_{\mathrm{freq}}(t)
=
\lambda_{\mathrm{LF}}^{(s)}
\mathcal{L}_{\mathrm{LF}}
+
\lambda_{\mathrm{HF}}^{(s)}
\mathcal{L}_{\mathrm{HF}},
\qquad
t\in\mathcal{S}_{s},
\label{eq:frequency_schedule}
\end{equation}
where $\mathcal{S}_{s}$ denotes the $s$-th optimization stage, and
$\lambda_{\mathrm{LF}}^{(s)}$ and $\lambda_{\mathrm{HF}}^{(s)}$
are fixed within that stage.
Across successive stages, the optimization emphasis shifts from low-frequency stabilization toward high-frequency refinement.
This staged frequency progression separates the formation of stable coarse structures from the recovery of fine spatial details, while avoiding competition between frequency objectives with different optimization roles.

\subsection{Canonical Group Consensus}
\label{sec:cgc}

Image-domain reconstruction objectives supervise each view independently and therefore provide only limited control over whether the same local structure is reconstructed consistently across viewpoints.
A straightforward cross-view comparison is also unreliable, since corresponding image locations may undergo substantial appearance changes caused by viewpoint variation, partial occlusion, and local deformation.
To address this issue, we introduce Canonical Group Consensus (CGC), which establishes cross-view supervision through a shared structural reference rather than direct image-space correspondence.
CGC first associates synchronized observations through canonical structural identities, then removes unreliable observations according to their geometric and visibility conditions, and finally consolidates the remaining observations into a feature-level consensus target.
In this way, cross-view information is introduced only after both structural correspondence and observation reliability have been established.

\noindent\textbf{Canonical Group Association.}
Let $\mathcal{M}_{c}$ denote the canonical facial surface introduced in Sec.~\ref{sec:preliminary}, whose topology provides persistent local structural identities across facial deformation.
We index a local canonical unit by its face identity $f$.
For frame $t$, the corresponding triangle on the refined posed surface
$\mathcal{M}_{t}$ is defined by vertices
$\mathbf{v}_{f,1}^{t}$,
$\mathbf{v}_{f,2}^{t}$, and
$\mathbf{v}_{f,3}^{t}$.
Its posed structural representative is given by
\begin{equation}
\mathbf{c}_{f}^{t}
=
\frac{1}{3}
\left(
\mathbf{v}_{f,1}^{t}
+
\mathbf{v}_{f,2}^{t}
+
\mathbf{v}_{f,3}^{t}
\right),
\label{eq:cgc_center}
\end{equation}
with the associated surface normal
\begin{equation}
\mathbf{n}_{f}^{t}
=
\operatorname{norm}
\left[
\left(
\mathbf{v}_{f,2}^{t}
-
\mathbf{v}_{f,1}^{t}
\right)
\times
\left(
\mathbf{v}_{f,3}^{t}
-
\mathbf{v}_{f,1}^{t}
\right)
\right],
\label{eq:cgc_normal}
\end{equation}
where $\operatorname{norm}(\cdot)$ denotes vector normalization.

For two synchronized views $A$ and $B$ observing the same frame $t$,
the same canonical identity $f$ determines a common local structure, while
$\mathbf{c}_{f}^{t}$ is projected into the two image planes according to their respective cameras:
\begin{equation}
\mathbf{p}_{f}^{A}
=
\pi
\left(
\mathbf{c}_{f}^{t};
\mathbf{\Pi}_{t}^{A}
\right),
\qquad
\mathbf{p}_{f}^{B}
=
\pi
\left(
\mathbf{c}_{f}^{t};
\mathbf{\Pi}_{t}^{B}
\right),
\label{eq:cgc_projection}
\end{equation}
where $\pi(\cdot)$ denotes perspective projection.
This construction avoids establishing correspondence directly between image pixels:
the correspondence is first defined by the persistent canonical identity and is only then transferred to the posed image observations.

We refer to the synchronized local observations associated with the same canonical identity as a \emph{canonical group}.
Importantly, this definition concerns the structural association rather than a particular primitive organization.
Its concrete realization can therefore follow the underlying avatar representation, while the subsequent consensus formulation remains unchanged.

\noindent\textbf{Reliability-Aware Observation Selection.}
Canonical association alone does not guarantee valid cross-view supervision.
A structurally corresponding location may be weakly visible, partially outside the foreground, or observed from an unfavorable surface orientation.
Directly enforcing consistency on such observations can introduce contradictory supervision.
We therefore evaluate the reliability of each canonical group before constructing the consensus target.

For view $v\in\{A,B\}$, let
$\mathbf{p}_{f}^{v}$ denote the projected location of group $f$,
and let $\mathbf{d}_{f}^{v}$ be the viewing direction from
$\mathbf{c}_{f}^{t}$ toward the camera center.
A group is considered reliable in view $v$ only when its projected representative lies inside the valid image domain, its surrounding patch is sufficiently supported by the foreground region, and the underlying surface is oriented toward the camera.
We express this validity as
\begin{equation}
r_{f}^{v}
=
\mathbb{I}_{\mathrm{proj}}
\,
\mathbb{I}
\left[
\rho
\left(
\mathbf{p}_{f}^{v}
\right)
\geq
\tau_{\mathrm{cov}}
\right]
\,
\mathbb{I}
\left[
\left\langle
\mathbf{n}_{f}^{t},
\mathbf{d}_{f}^{v}
\right\rangle
>
\tau_{\mathrm{n}}
\right],
\label{eq:cgc_reliability}
\end{equation}
where
$\mathbb{I}_{\mathrm{proj}}$
checks positive depth and valid image projection,
$\rho(\mathbf{p}_{f}^{v})$
measures the foreground coverage within a local patch centered at
$\mathbf{p}_{f}^{v}$,
and
$\tau_{\mathrm{cov}}$ and $\tau_{\mathrm{n}}$
are the coverage and orientation thresholds, respectively.

Only groups satisfying
\begin{equation}
r_{f}^{A} r_{f}^{B}=1
\label{eq:cgc_common_valid}
\end{equation}
are retained for cross-view supervision.
This intersection is important because the existence of a canonical correspondence does not imply that both observations contain equally trustworthy visual evidence.
The reliability filtering therefore separates \emph{structural correspondence} from \emph{observation usability}, preventing occluded or weakly supported regions from contaminating the consensus. When depth observations are available, we additionally reject projected groups whose estimated depth is inconsistent with the observed depth beyond a predefined tolerance, further excluding occluded observations from consensus construction.

The auxiliary view is sampled from synchronized cameras with a sufficient viewpoint separation from the current training view.
This provides complementary visual evidence while avoiding redundant near-identical observations or excessively separated viewpoints with weak overlap.

\noindent\textbf{Feature-Level Consensus Supervision.}
For each retained canonical group $f$, we extract three local observations:
a rendered patch from the current view,
a ground-truth patch from the same view,
and a ground-truth patch from the synchronized auxiliary view:
\begin{equation}
\mathbf{P}_{f}^{r}
=\mathcal{C}_{\mathbf{p}_{f}^{A}}(\hat{\mathbf{I}}_{t}^{A}),
\quad
\mathbf{P}_{f}^{A}
=\mathcal{C}_{\mathbf{p}_{f}^{A}}(\mathbf{I}_{t}^{A}),
\quad
\mathbf{P}_{f}^{B}
=\mathcal{C}_{\mathbf{p}_{f}^{B}}(\mathbf{I}_{t}^{B}).
\label{eq:cgc_patches}
\end{equation}
where $\mathcal{C}_{p}(\cdot)$ denotes local patch extraction centered at image location $p$. Because $P_f^A$ and $P_f^B$ originate from the same canonical structure at the same facial state, they provide
complementary observations of the same local content.

\begin{figure*}[t]
  \centering
  
  \includegraphics[width=\textwidth]{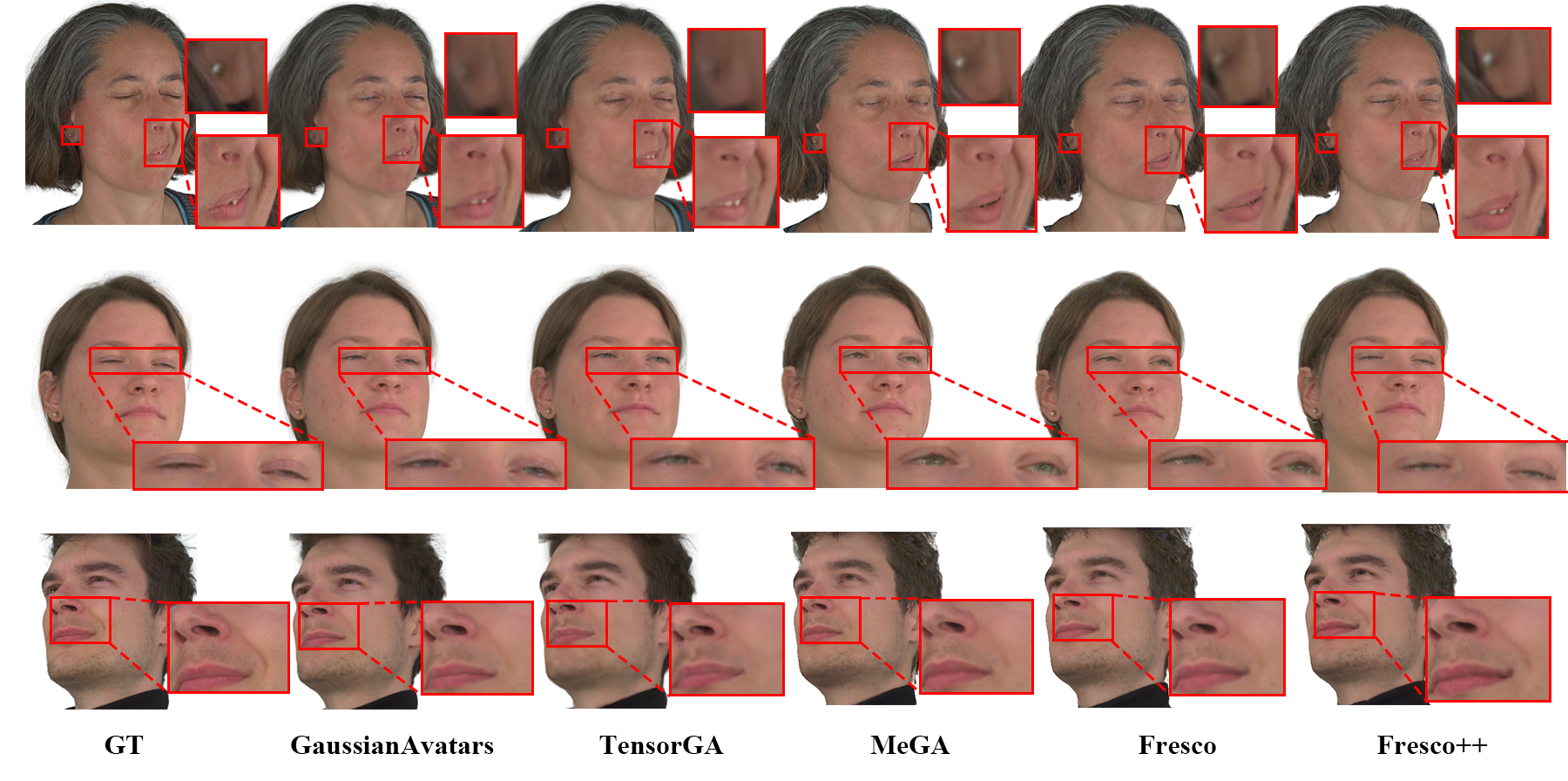}

   \caption{Qualitative comparison on self-reenactment.
Fresco++ better preserves subject-specific appearance details and reproduces expression-dependent local structures, such as earrings, eye opening, and cheek wrinkles, under unseen facial motions.}
   \label{fig:self_reenactment}
\end{figure*}

Rather than enforcing pixel-wise agreement across views, we map these patches into a perceptual feature space using a frozen feature extractor $\Phi$:
\begin{equation}
\mathbf{f}_{f}^{r}
=
\Phi
\left(
\mathbf{P}_{f}^{r}
\right),
\qquad
\mathbf{f}_{f}^{A}
=
\Phi
\left(
\mathbf{P}_{f}^{A}
\right),
\qquad
\mathbf{f}_{f}^{B}
=
\Phi
\left(
\mathbf{P}_{f}^{B}
\right).
\label{eq:cgc_features}
\end{equation}
We then construct a cross-view consensus representation from the two ground-truth observations:
\begin{equation}
\bar{\mathbf{f}}_{f}
=
\operatorname{sg}
\left[
(1-\alpha)\mathbf{f}_{f}^{A}
+
\alpha\mathbf{f}_{f}^{B}
\right],
\label{eq:cgc_consensus}
\end{equation}
where $\alpha$ controls the contribution of the auxiliary observation and
$\operatorname{sg}(\cdot)$ denotes stop-gradient.

The CGC objective is defined over the set
$\mathcal{V}_{t}$ of reliable canonical groups:
\begin{equation}
\mathcal{L}_{\mathrm{CGC}}
=
\frac{1}{|\mathcal{V}_{t}|}
\sum_{f\in\mathcal{V}_{t}}
\left\|
\mathbf{f}_{f}^{r}
-
\bar{\mathbf{f}}_{f}
\right\|_{1}.
\label{eq:cgc_loss}
\end{equation}
Only the rendered branch remains differentiable, while the ground-truth consensus is treated as a fixed supervisory target.

This asymmetric formulation differs fundamentally from directly enforcing agreement between two rendered views.
CGC does not require an additional rendering of the auxiliary viewpoint and does not propagate errors from one prediction into another.
Instead, synchronized ground-truth observations are first consolidated into a reliable local reference, and the current rendering is optimized toward this consensus.
Consequently, the supervision reflects visual evidence that is shared across views while remaining anchored to the current reconstruction objective.

\subsection{Joint Optimization}
\label{sec:joint_optimization}

Given a training observation $\mathbf{I}_{t}^{v}$ and the corresponding rendered image
$\hat{\mathbf{I}}_{t}^{v}$, the avatar parameters are optimized under three complementary sources of supervision.
The reconstruction objective $\mathcal{L}_{\mathrm{rec}}$ measures the fidelity of the rendered avatar to the observed image and regularizes its geometry during fitting.
The proposed frequency objective $\mathcal{L}_{\mathrm{freq}}$ controls the coarse-to-fine evolution of image details, while
$\mathcal{L}_{\mathrm{CGC}}$ introduces reliable cross-view constraints on corresponding local structures.
The complete objective is
\begin{equation}
\mathcal{L}_{\mathrm{total}}
=
\mathcal{L}_{\mathrm{rec}}
+
\mathcal{L}_{\mathrm{freq}}
+
\lambda_{\mathrm{CGC}}
\mathcal{L}_{\mathrm{CGC}},
\label{eq:total_loss}
\end{equation}
where $\mathcal{L}_{\mathrm{freq}}$ follows the stage-wise curriculum defined in Eq.~\eqref{eq:frequency_schedule}, and
$\lambda_{\mathrm{CGC}}$ controls the strength of cross-view consensus supervision.

These objectives play different roles during reconstruction.
$\mathcal{L}_{\mathrm{rec}}$ provides direct per-view fitting between the rendered avatar and the captured observations.
$\mathcal{L}_{\mathrm{freq}}$ further organizes this fitting process across spatial frequencies, first emphasizing stable low-frequency structures and subsequently refining high-frequency details.
In contrast, $\mathcal{L}_{\mathrm{CGC}}$ couples synchronized views through canonical structural correspondence and constrains each local region using consensus evidence collected from multiple observations.
Together, the three objectives promote accurate per-view reconstruction, progressive detail recovery, and cross-view structural consistency within a unified optimization process.

\begin{figure*}[t]
  \centering
  
  \includegraphics[width=\textwidth]{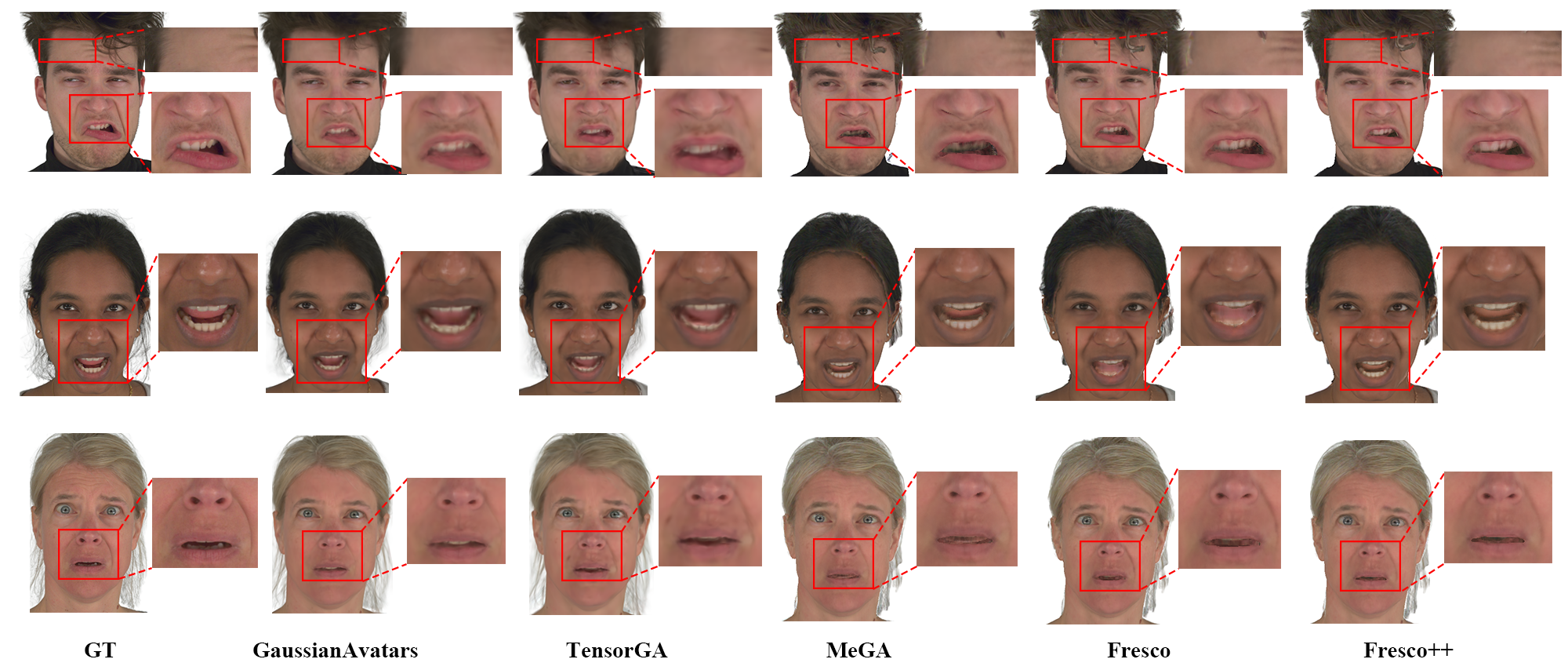}

   \caption{Qualitative comparison on novel-view synthesis.
Fresco++ produces more faithful fine-grained details under unseen viewpoints, particularly in challenging regions such as teeth and facial wrinkles, while maintaining more stable local appearance across view changes.}
   \label{fig:novel_view}
\end{figure*}
\section{Experiments}
\label{sec:experiments}

We conduct extensive experiments to evaluate Fresco++ from four complementary perspectives.
First, we compare Fresco++ with representative head-avatar methods on novel-view synthesis and reenactment to assess its overall reconstruction and animation quality.
Second, we perform controlled ablations to examine the contribution of individual components and to compare the proposed Canonical Group Consensus with the UV-space consistency used in our preliminary formulation.
Third, we investigate different low- and high-frequency optimization orders to analyze the effect of the stage-wise frequency curriculum.
Finally, we transfer the proposed strategy to different avatar representations to examine its representation transferability.

\noindent\textbf{Dataset and Preprocessing.}
We conduct experiments on the NeRSemble dataset~\cite{kirschstein2023nersemble}, which contains synchronized multi-view videos of multiple subjects captured in a controlled studio using 16 calibrated cameras.
Following the evaluation protocols of GaussianAvatars~\cite{qian2024gaussianavatars} and MeGA~\cite{wang2025mega}, we select eight subjects for evaluation.
Each subject contains eleven sequences, including ten sequences with predefined expressions or emotions and one free-form sequence.

Following prior preprocessing pipelines~\cite{qian2024gaussianavatars,wang2025mega}, all frames are resized to a resolution of $802\times550$, and foreground masks are generated for each view.
We additionally obtain semantic facial parsing maps using~\cite{lee2020maskgan} and reconstruct per-frame depth maps using Metashape~\cite{over2021processing}.
Unless otherwise specified, we use the same training and evaluation split across all compared methods.
For each subject, nine predefined sequences and 15 camera views are used for training, while one predefined sequence and the remaining camera view are held out for evaluation.

\noindent\textbf{Evaluation Settings.}
We evaluate the reconstructed avatars under three settings: novel-view synthesis, self-reenactment, and cross-reenactment. For \emph{novel-view synthesis}, we evaluate generalization to unseen viewpoints by rendering the avatar from the camera view excluded from training and comparing the results with the corresponding ground-truth images. For \emph{self-reenactment}, we evaluate generalization to unseen facial motions by driving each avatar with the poses and expressions from the held-out sequence of the same subject and rendering the resulting animation across all sixteen camera views. For \emph{cross-reenactment}, we further evaluate cross-identity motion transfer by driving a target avatar with poses and expressions extracted from another subject and rendering the results under the target subject's camera configuration.
Since paired ground-truth images are unavailable in this setting, cross-reenactment is evaluated qualitatively.

\begin{table}[t]
  \centering
  \caption{Quantitative comparison with previous methods on novel-view synthesis and self-reenactment. Best results are \textbf{bold}, and second-best results are \underline{underlined}.}
  \label{tab:mega_comp}
  \setlength{\tabcolsep}{3.8pt}
  \renewcommand{\arraystretch}{1.05}
  \resizebox{\columnwidth}{!}{%
  \begin{tabular}{lccc|ccc}
    \toprule
    \multirow{2}{*}{Method}
    & \multicolumn{3}{c}{\textbf{Novel-View}}
    & \multicolumn{3}{c}{\textbf{Self-Reenactment}} \\
    \cmidrule(lr){2-4} \cmidrule(lr){5-7}
    & PSNR$\uparrow$ & SSIM$\uparrow$ & LPIPS$\downarrow$
    & PSNR$\uparrow$ & SSIM$\uparrow$ & LPIPS$\downarrow$ \\
    \midrule

    Gaussian Head Avatar
    & 29.84 & 0.895 & 0.083
    & 22.34 & 0.851 & 0.145 \\

    EMavatar
    & 30.16 & 0.898 & 0.082
    & 22.56 & 0.852 & 0.145 \\

    GaussianAvatars
    & 31.50 & 0.935 & 0.060
    & 29.75 & \underline{0.930} & 0.066 \\

    MeGA
    & 30.79 & 0.932 & 0.066
    & 29.62 & 0.926 & 0.071 \\

    TensorGA
    & 31.28 & 0.931 & 0.067
    & 29.90 & \textbf{0.931} & 0.071 \\

    \midrule

    Fresco
    & \underline{31.80} & \underline{0.938} & \underline{0.058}
    & \underline{30.30} & 0.929 & \underline{0.064} \\

    Fresco++ (Ours)
    & \textbf{32.18} & \textbf{0.939} & \textbf{0.056}
    & \textbf{30.61} & \textbf{0.931} & \textbf{0.063} \\

    \bottomrule
  \end{tabular}}
\end{table}

\noindent\textbf{Evaluation Metrics.}
For novel-view synthesis and self-reenactment, we evaluate image fidelity using PSNR and SSIM, and perceptual similarity using LPIPS.
All metrics are computed within a consistent foreground evaluation region for all methods to avoid the influence of background pixels.
Higher PSNR and SSIM and lower LPIPS indicate better reconstruction quality.
Cross-reenactment is evaluated qualitatively because no paired ground-truth target is available.

\noindent\textbf{Implementation Details.}
All experiments are conducted on a single NVIDIA RTX A6000 GPU.
For MeGA, Fresco, and Fresco++, we train each subject for 100 epochs with a batch size of 16 and an initial learning rate of $1\times10^{-3}$.
GaussianAvatars and TensorGA are trained for 600K iterations following their official training protocols.
All methods use an input resolution of $802\times550$, and the optimization settings are kept fixed across subjects unless otherwise specified.

For Fresco++, low-frequency supervision and CGC are activated from epochs 10 to 50 with $\lambda_{\mathrm{LF}}=0.1$ and $\lambda_{\mathrm{CGC}}=0.002$, respectively, after an initial warm-up stage.
High-frequency supervision is subsequently introduced from epochs 50 to 80 with $\lambda_{\mathrm{HF}}=0.005$, forming a progressive low-to-high frequency curriculum.

For CGC, synchronized observations with relative viewing angles of $25^\circ$--$55^\circ$ are considered, with views around $40^\circ$ preferred.
Unreliable observations are further discarded according to foreground validity, surface visibility, and depth-based occlusion checks. We use a frozen VGG-16 encoder as the feature extractor $\Phi$ and set the consensus weight $\alpha=0.5$.
CGC is used only during training and introduces no additional inference-time computation.

\subsection{Comparisons with State-of-the-Art Methods}
\label{sec:sota_comparison}

We compare Fresco++ with representative state-of-the-art head avatar methods, including Gaussian Head Avatar~\cite{xu2024gaussian}, EMavatar~\cite{zhang2025eavatar}, GaussianAvatars~\cite{qian2024gaussianavatars}, MeGA~\cite{wang2025mega}, and TensorGA~\cite{wang20253d}, as well as Fresco~\cite{zhang2026fresco}.
All compared methods are evaluated under the same data split and evaluation settings described above.

\noindent\textbf{Quantitative Comparison.}
Table~\ref{tab:mega_comp} summarizes the quantitative comparison on novel-view synthesis and self-reenactment.
Overall, Fresco++ delivers the strongest performance across the two evaluation settings.
On novel-view synthesis, it achieves the best results on all three metrics, with a PSNR of 32.18 dB.
Compared with GaussianAvatars, a strong competing baseline in this setting, Fresco++ gains 0.68 dB in PSNR, indicating improved reconstruction fidelity when the avatar is observed from unseen viewpoints.

\begin{figure*}[t]
  \centering
  
  \includegraphics[width=0.9\textwidth]{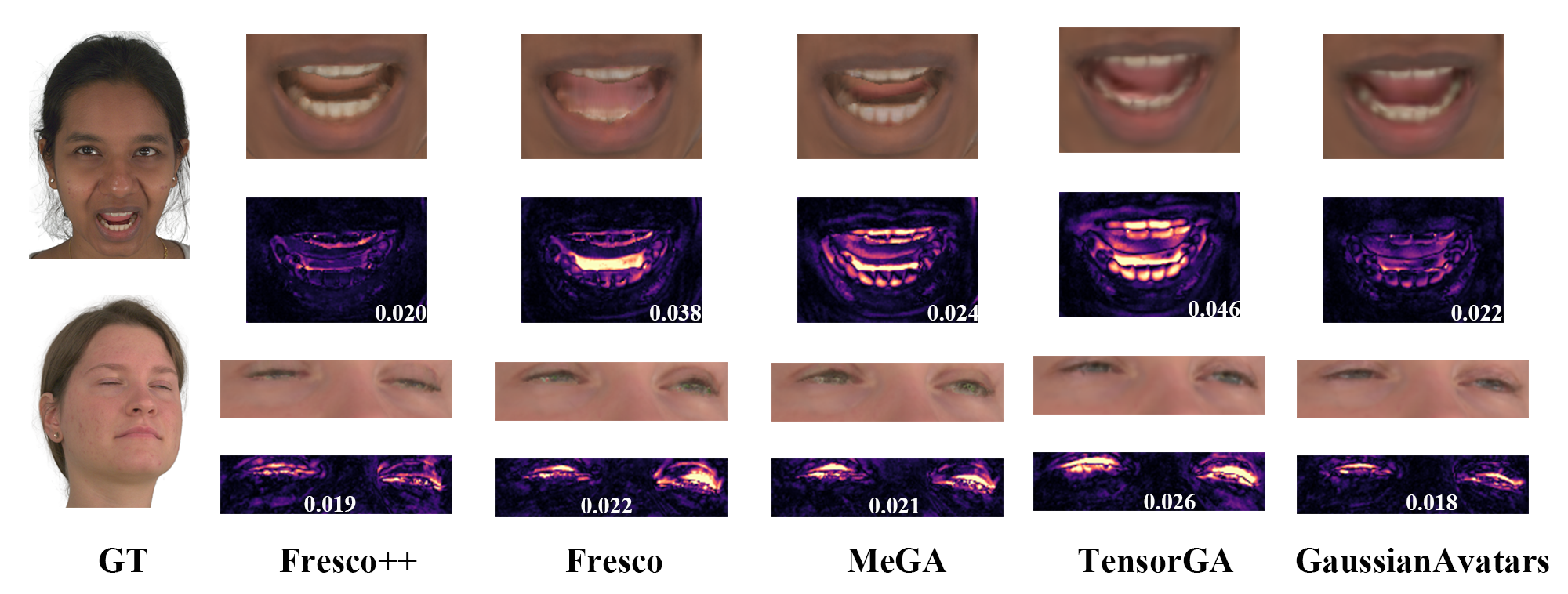}

   \caption{Local reconstruction error comparison on self-reenactment and novel-view synthesis.
For each example, we show enlarged RGB regions together with the corresponding error maps.
Fresco++ exhibits lower local reconstruction errors in challenging facial regions, particularly around the mouth and eyes.
The value shown in each error map denotes the mean reconstruction error within the displayed region, and all error maps are visualized using the same color scale.}
   \label{fig:error_map}
\end{figure*}

The same trend is observed in self-reenactment.
Fresco++ achieves the highest PSNR and the lowest LPIPS, while reaching the best SSIM jointly with TensorGA.
The consistent improvement over Fresco in both novel-view synthesis and self-reenactment further suggests that the proposed formulation benefits not only viewpoint generalization but also motion generalization.

\begin{figure*}[t]
  \centering
  
  \includegraphics[width=0.85\textwidth]{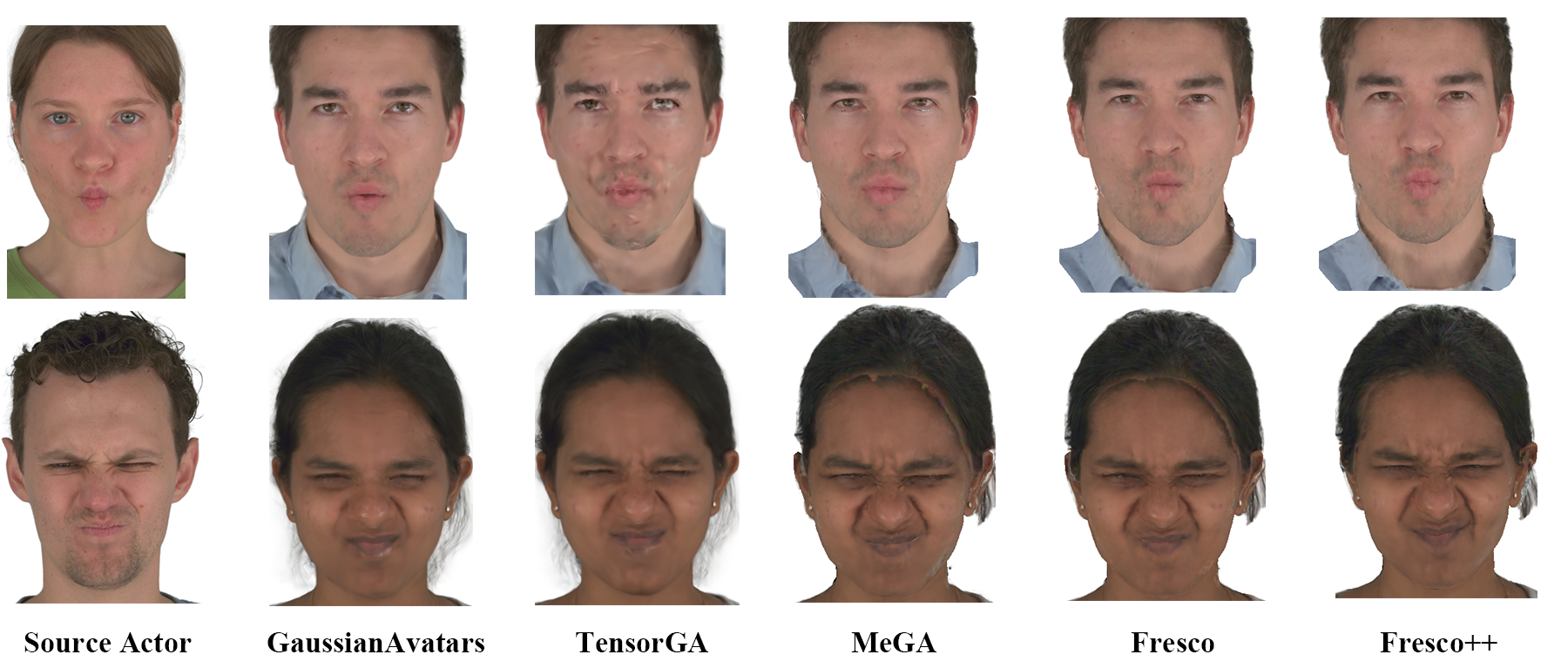}

   \caption{Qualitative comparison on cross-reenactment.
The source actor in the first column provides the driving expression for the target avatars.
Fresco++ reproduces the transferred facial motions more faithfully, including mouth deformation and strong eye-region contraction, while better preserving the target subject's local appearance.}
   \label{fig:cross_reenactment}
\end{figure*}

\noindent\textbf{Qualitative Comparison.}
The visual comparisons further demonstrate the advantages of Fresco++ in preserving fine-grained appearance details and expression-dependent local structures.

As shown in Fig.~\ref{fig:self_reenactment}, Fresco++ produces more faithful self-reenactment results across different facial motions.
Compared with competing methods, it better preserves small appearance details, such as earrings, while reproducing expression-dependent structures more accurately, including the degree of eye opening and fine cheek wrinkles.
These results indicate that the reconstructed avatar not only retains subject-specific visual details, but also responds more faithfully to unseen facial motions during reenactment.

Figure~\ref{fig:novel_view} presents the qualitative comparison on novel-view synthesis.
Fresco++ produces more realistic local details under unseen viewpoints, particularly in challenging regions such as teeth and facial wrinkles.
While competing methods may exhibit over-smoothed textures or unstable local appearance when the viewing direction changes, Fresco++ better preserves these fine structures and produces results that are visually closer to the ground truth.
This demonstrates improved view generalization and local reconstruction fidelity under unseen viewpoints.

To further highlight local differences that may be less obvious from rendered images alone, we visualize reconstruction errors in representative facial regions for both self-reenactment and novel-view synthesis.
As shown in Fig.~\ref{fig:error_map}, Fresco++ produces lower local errors in challenging expression-sensitive regions, especially around the mouth and eyes.
These error maps further support the qualitative observations above, showing that Fresco++ better preserves local structures and appearance details under both unseen facial motions and unseen viewpoints.

Figure~\ref{fig:cross_reenactment} further evaluates cross-reenactment, where the target avatar is driven by expressions from another subject.
Fresco++ better preserves the target subject's identity-specific appearance while transferring the driving expressions more faithfully.
Compared with competing methods, it maintains more stable local structures around the mouth and eyes and better preserves fine appearance details under cross-identity motion transfer.
These results suggest that Fresco++ achieves a better balance between identity preservation and expression transfer in cross-reenactment.

\begin{figure*}[t]
  \centering
  
  \includegraphics[width=\textwidth]{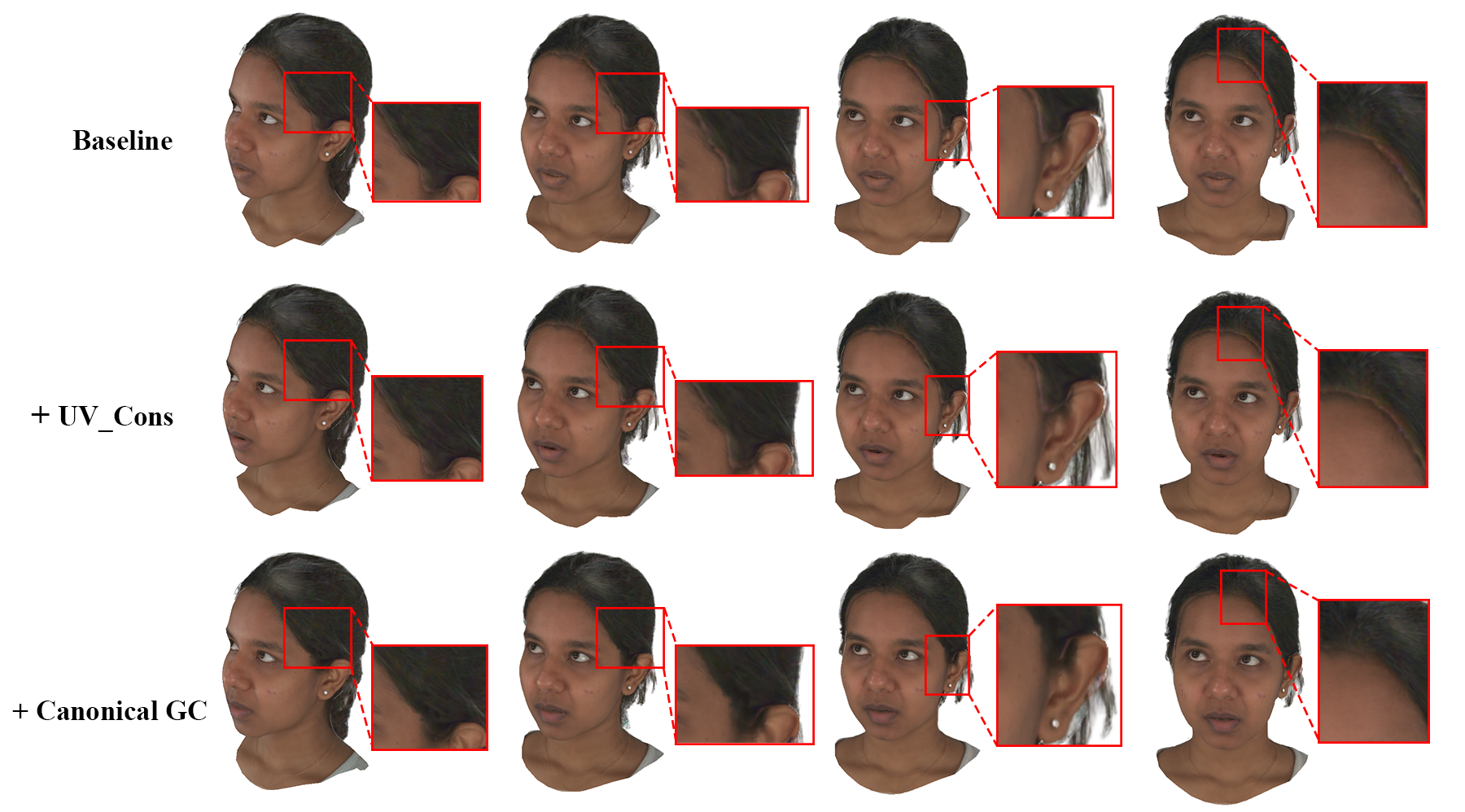}

   \caption{Qualitative ablation of cross-view consistency.
We compare the baseline, the UV-space consistency used in Fresco, and the proposed CGC across multiple viewpoints.
The enlarged regions show that CGC preserves more stable local structures and appearance under viewpoint changes, with clearer boundaries and more consistent fine-grained details.}
   \label{fig:uv_cgc_ablation}
\end{figure*}

\subsection{Ablation Studies}
\label{sec:ablation}

We conduct the ablation studies using the hybrid mesh--Gaussian representation adopted in our primary implementation.
Unless otherwise specified, ``Baseline'' refers to its standard optimization pipeline without UV-space consistency, CGC, or the proposed frequency curriculum.

\begin{table}[t]
  \centering
  \caption{
  Ablation study of the main components in Fresco++.
  The reported results are averaged over all evaluated subjects under novel-view synthesis and self-reenactment.
  UV Cons. denotes the UV-space consistency component used in Fresco.
  Best results are \textbf{bold}, and second-best results are \underline{underlined}.
  }
  \label{tab:ablation}
  \setlength{\tabcolsep}{3.8pt}
  \renewcommand{\arraystretch}{1.05}
  \resizebox{\columnwidth}{!}{%
  \begin{tabular}{lccc|ccc}
    \toprule
    \multirow{2}{*}{Variant}
    & \multicolumn{3}{c}{\textbf{Novel-View}}
    & \multicolumn{3}{c}{\textbf{Self-Reenactment}} \\
    \cmidrule(lr){2-4} \cmidrule(lr){5-7}
    & PSNR$\uparrow$ & SSIM$\uparrow$ & LPIPS$\downarrow$
    & PSNR$\uparrow$ & SSIM$\uparrow$ & LPIPS$\downarrow$ \\
    \midrule

    Baseline
    & 30.79 & 0.932 & 0.066
    & 29.62 & 0.926 & 0.071 \\

    + UV Cons.
    & 31.17 & 0.933 & 0.064
    & 29.88 & 0.927 & 0.070 \\

    + CGC
    & \underline{31.50} & \underline{0.935} & \underline{0.060}
    & \underline{30.06} & \underline{0.928} & \underline{0.067} \\

    Fresco++
    & \textbf{32.18} & \textbf{0.939} & \textbf{0.056}
    & \textbf{30.61} & \textbf{0.931} & \textbf{0.063} \\

    \bottomrule
  \end{tabular}}
\end{table}

\noindent\textbf{Component Analysis.}
Table~\ref{tab:ablation} reports the average results over all evaluated subjects.
In addition to the proposed CGC, we include the UV-space consistency component used in Fresco as a reference to directly compare the two cross-view consistency formulations under the same baseline.
Starting from this common baseline, both UV-space consistency and CGC improve reconstruction quality under novel-view synthesis and self-reenactment.
However, CGC consistently outperforms the UV-based variant across all evaluation metrics.
For novel-view synthesis, CGC improves the average PSNR from 30.79 dB to 31.50 dB and reduces LPIPS from 0.066 to 0.060, whereas the UV-space consistency variant reaches 31.17 dB and 0.064, respectively.
A similar trend is observed in self-reenactment, where CGC further improves reconstruction fidelity and perceptual quality over the UV-based formulation.
These results suggest that associating multi-view observations through canonical structural correspondence provides more effective cross-view supervision than consistency established in UV space.

Combining CGC with the proposed frequency curriculum further improves the reconstruction quality, leading to the complete Fresco++ model.
Fresco++ achieves the best performance across all metrics under both evaluation settings, reaching 32.18 dB PSNR and 0.056 LPIPS on novel-view synthesis, and 30.61 dB PSNR and 0.063 LPIPS on self-reenactment.
The additional improvement over the CGC-only variant indicates that canonical cross-view consistency and coarse-to-fine frequency optimization provide complementary benefits.
While CGC mainly improves the consistency of corresponding local structures across viewpoints, the frequency curriculum further facilitates the recovery of fine-grained visual details.

\noindent\textbf{Qualitative Analysis of Cross-View Consistency.}
We further compare the baseline, the UV-space consistency used in Fresco, and the proposed CGC across multiple viewpoints in Fig.~\ref{fig:uv_cgc_ablation}.
As the viewing direction changes, the baseline exhibits noticeable instability in local appearance and structural boundaries, particularly around the hairline, temporal region, and ear boundary.
Introducing UV-space consistency alleviates these cross-view discrepancies to some extent, but local details may still vary across viewpoints.
In contrast, CGC maintains more stable local structures and appearance across different views, while preserving clearer boundaries and more consistent fine-grained details.

These qualitative observations are consistent with the quantitative results in Table~\ref{tab:ablation}.
Although the UV-based formulation already improves the baseline, CGC establishes cross-view structural associations through shared canonical structural identities and constructs consistency supervision from reliable multi-view observations, thereby providing more effective cross-view constraints.
The resulting improvements are reflected not only in the reconstruction quality of individual views, but also in the stability of corresponding local structures under viewpoint changes.

\subsection{Analysis of Frequency Curriculum}
\label{sec:frequency_analysis}

\noindent\textbf{Effect of Frequency Scheduling.}
We further examine the effect of frequency ordering on Subject 306 by comparing three training schedules under otherwise identical settings.
All variants are trained for the same number of epochs with identical optimization settings, while CGC and other additional consistency constraints are disabled.
The only difference lies in the frequency schedule: LF$\rightarrow$HF first applies low-frequency supervision and then switches to high-frequency supervision; HF$\rightarrow$LF reverses this order; and LF+HF activates both low- and high-frequency objectives throughout the corresponding frequency-guided stages.
All variants subsequently undergo the same final stage without frequency supervision.

As shown in Table~\ref{tab:frequency_schedule}, staged frequency optimization consistently outperforms simultaneous low- and high-frequency supervision.
Among the two staged strategies, LF$\rightarrow$HF achieves the strongest overall performance, yielding the best results on self-reenactment and the highest PSNR and SSIM on novel-view synthesis.
These results indicate that the order of frequency supervision matters, and that prioritizing low-frequency guidance before high-frequency refinement leads to a more effective coarse-to-fine optimization process.

\begin{table}[t]
\centering
\caption{
Comparison of different frequency optimization schedules on Subject 306.
All variants are trained under the same settings and differ only in the scheduling of low- and high-frequency supervision.
Best results are \textbf{bold}, and second-best results are \underline{underlined}.
}
\label{tab:frequency_schedule}
\resizebox{\columnwidth}{!}{
\begin{tabular}{lccc|ccc}
\hline
\multirow{2}{*}{Schedule}
& \multicolumn{3}{c|}{Novel-View Synthesis}
& \multicolumn{3}{c}{Self-Reenactment} \\
& PSNR$\uparrow$ & SSIM$\uparrow$ & LPIPS$\downarrow$
& PSNR$\uparrow$ & SSIM$\uparrow$ & LPIPS$\downarrow$ \\
\hline
LF+HF
& 31.92
& 0.960
& 0.039
& 30.83
& 0.952
& 0.046 \\

HF$\rightarrow$LF
& \underline{32.25}
& \underline{0.961}
& \textbf{0.037}
& \underline{31.17}
& \underline{0.953}
& \underline{0.045} \\

LF$\rightarrow$HF
& \textbf{32.37}
& \textbf{0.962}
& \textbf{0.037}
& \textbf{31.34}
& \textbf{0.955}
& \textbf{0.044} \\
\hline
\end{tabular}
}
\end{table}

\noindent\textbf{Optimization Stability.}
We further compare the optimization stability of the three frequency schedules using the cumulative variation of the training-time head RGB and SSIM losses.
A lower value indicates a smoother reconstruction trajectory with smaller loss fluctuations during training.

\begin{figure}[t]
\centering
  
\includegraphics[width=0.95\linewidth]{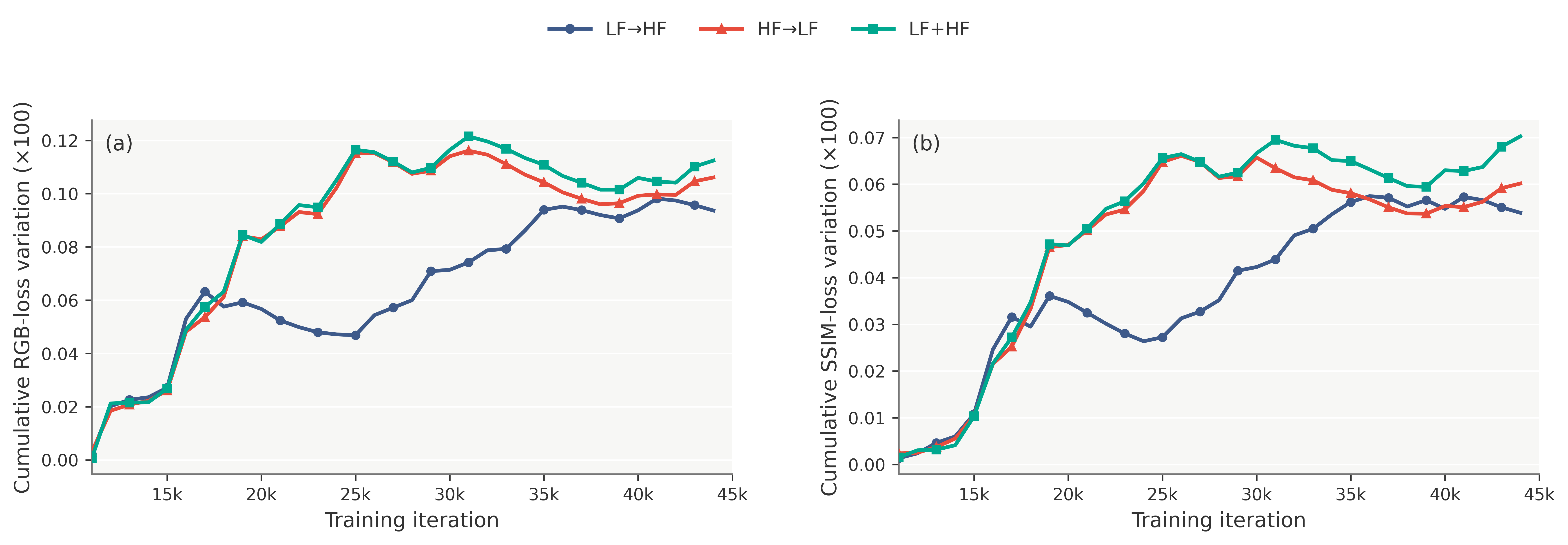}

\caption{Optimization stability under different frequency schedules on Subject 306.
We report the cumulative variation of the training-time head RGB loss in (a) and head SSIM loss in (b).
Lower values indicate smaller changes in the reconstruction trajectory. After the early stage of training, LF$\rightarrow$HF exhibits lower cumulative variation than HF$\rightarrow$LF and simultaneous low- and high-frequency optimization.}
   \label{fig:frequency_stability}
\end{figure}

As shown in Fig.~\ref{fig:frequency_stability}, LF$\rightarrow$HF exhibits consistently lower cumulative variation in both loss terms after the early stage of training.
In contrast, HF$\rightarrow$LF and simultaneous low- and high-frequency optimization accumulate larger variations as training proceeds.
This suggests that prioritizing low-frequency supervision first can reduce subsequent fluctuations in the reconstruction trajectory, providing a more stable transition toward high-frequency refinement.

\noindent\textbf{Effect of Early Frequency Supervision.}
We further examine the effect of frequency supervision during the early stage of optimization on Subject 074 by visualizing the reconstruction results at the 30th epoch.
Under otherwise identical training settings, we compare applying high-frequency supervision (Early HF) and low-frequency supervision (Early LF) at this stage.

\begin{figure}[t]
\centering
  
\includegraphics[width=0.95\linewidth]{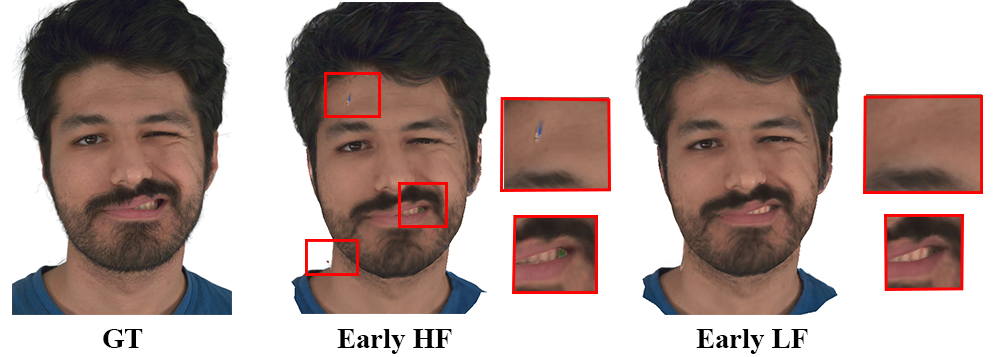}

\caption{Qualitative comparison of different frequency supervision strategies at the early stage of optimization.
Results are shown at the 30th training epoch.
Early HF tends to introduce spurious local high-frequency details, whereas Early LF produces cleaner and more coherent intermediate reconstructions.}
   \label{fig:early_frequency}
\end{figure}

As shown in Fig.~\ref{fig:early_frequency}, applying high-frequency supervision at the early stage tends to amplify small local intensity variations into spurious high-frequency details, particularly around the forehead and mouth regions.
In contrast, low-frequency supervision produces cleaner and more coherent intermediate reconstructions, with fewer premature local artifacts.
These results indicate that the choice of frequency supervision at the early optimization stage directly affects the quality of the intermediate reconstruction, and that emphasizing low-frequency information first helps avoid premature high-frequency artifacts.

\subsection{Transferability across Avatar Representations}
\label{sec:transferability}

To evaluate the generality and transferability of the proposed optimization strategy across different avatar representations, we further integrate Fresco++ into two representative Gaussian-based avatar models, GaussianAvatars and TensorGA.
Unlike the hybrid mesh--Gaussian representation adopted in our primary implementation, both models rely on purely Gaussian representations, providing a distinct setting for examining whether the proposed frequency curriculum and CGC remain effective beyond the original representation.

The frequency curriculum can be naturally applied to the rendered outputs without modifying the underlying representation.
For CGC, we retain the same canonical consensus formulation while adapting the construction of canonical groups according to the structural associations available in each representation.
We then compare the original and Fresco++-enhanced models under novel-view synthesis and self-reenactment, together with qualitative results and computational overhead.

\begin{figure}[t]
\centering
  
\includegraphics[width=0.95\linewidth]{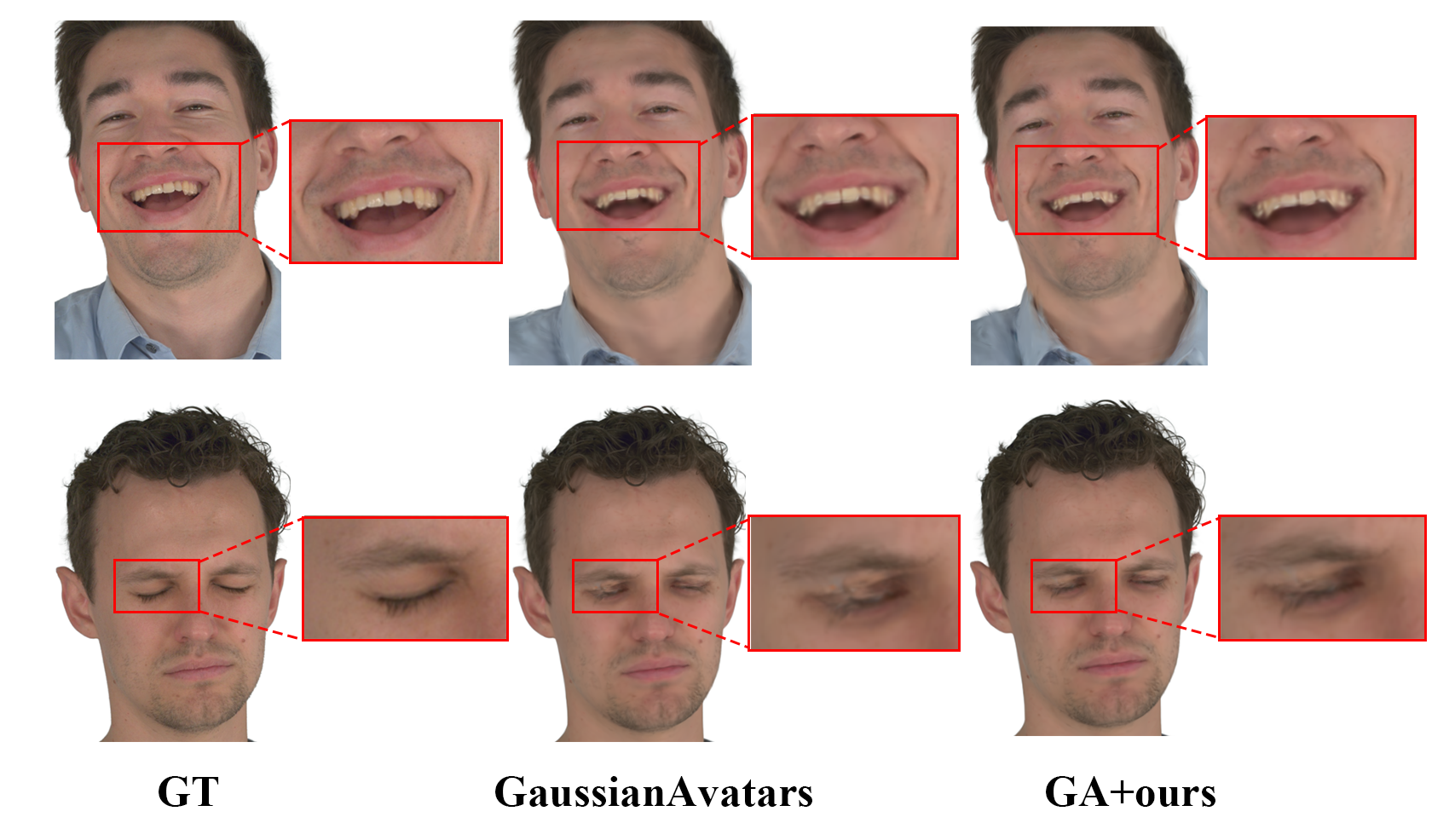}

\caption{Qualitative comparison before and after transferring Fresco++ to GaussianAvatars.
The enhanced model better recovers expression-dependent local details, including mouth-corner wrinkles and surrounding facial textures, while reducing local artifacts around the eye region.}
   \label{fig:ga_transfer}
\end{figure}

\noindent\textbf{Quantitative Evaluation.}
Table~\ref{tab:transferability} reports the quantitative results after integrating Fresco++ into GaussianAvatars and TensorGA.
Both models show consistent gains under novel-view synthesis, with improved PSNR and SSIM and reduced LPIPS.
A similar trend is observed for self-reenactment, where PSNR and LPIPS are further improved while SSIM remains stable.
The consistent improvements across the two pure Gaussian representations further verify the transferability of the proposed optimization strategy across different avatar representations.

\begin{table}[t]
\centering
\caption{
Quantitative evaluation of Fresco++ on different Gaussian avatar representations.
We compare GaussianAvatars and TensorGA before and after integrating the proposed frequency curriculum and CGC under novel-view synthesis and self-reenactment.
}
\label{tab:transferability}
\resizebox{\columnwidth}{!}{
\begin{tabular}{llccc|ccc}
\hline
\multirow{2}{*}{Representation}
& \multirow{2}{*}{Variant}
& \multicolumn{3}{c|}{Novel-View Synthesis}
& \multicolumn{3}{c}{Self-Reenactment} \\
& & PSNR$\uparrow$ & SSIM$\uparrow$ & LPIPS$\downarrow$
& PSNR$\uparrow$ & SSIM$\uparrow$ & LPIPS$\downarrow$ \\
\hline

\multirow{2}{*}{GaussianAvatars}
& Original
& 31.50
& 0.935
& 0.060
& 29.75
& \textbf{0.930}
& 0.066 \\

& + Fresco++
& \textbf{31.92}
& \textbf{0.936}
& \textbf{0.058}
& \textbf{29.86}
& \textbf{0.930}
& \textbf{0.065} \\
\hline

\multirow{2}{*}{TensorGA}
& Original
& 31.28
& 0.931
& 0.067
& 29.90
& \textbf{0.931}
& 0.071 \\

& + Fresco++
& \textbf{31.61}
& \textbf{0.933}
& \textbf{0.066}
& \textbf{30.03}
& \textbf{0.931}
& \textbf{0.069} \\
\hline
\end{tabular}
}
\end{table}

\begin{figure}[t]
\centering
  
\includegraphics[width=0.95\linewidth]{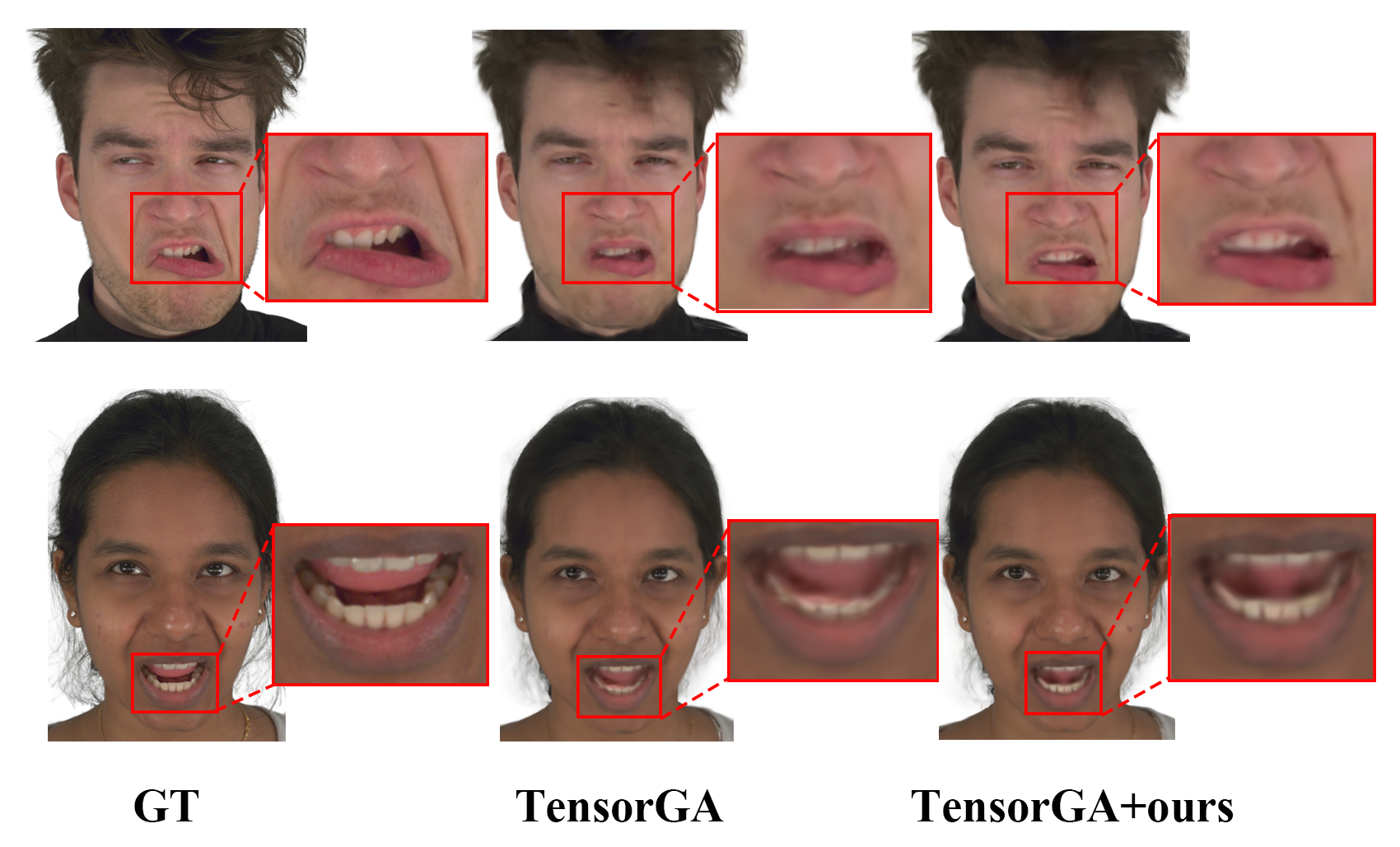}

\caption{Qualitative comparison before and after transferring Fresco++ to TensorGA.
The enhanced model produces clearer local mouth textures and more accurately reconstructs the teeth structure.}
   \label{fig:tensorga_transfer}
\end{figure}

\noindent\textbf{Qualitative Evaluation.}
On GaussianAvatars, the qualitative comparison in Fig.~\ref{fig:ga_transfer} shows that incorporating Fresco++ improves the reconstruction of expression-dependent local appearance.
In particular, the mouth-corner wrinkles and surrounding facial textures are recovered more faithfully and more closely match the ground truth.
The eye region also exhibits cleaner reconstruction with fewer local artifacts. A similar improvement can be observed on TensorGA in Fig.~\ref{fig:tensorga_transfer}.
After integrating Fresco++, local textures around the mouth become clearer and more faithful, while the teeth structure is reconstructed more accurately.
These qualitative observations are consistent with the quantitative gains reported in Table~\ref{tab:transferability}.

\noindent\textbf{Computational Overhead.}
We further evaluate the computational overhead introduced by transferring Fresco++ to GaussianAvatars and TensorGA.
The comparison is conducted on two representative subjects under the same hardware and training settings, and we report the average training time, peak GPU memory usage, and inference time.

\begin{table}[t]
\centering
\caption{
Computational overhead of transferring Fresco++ to different Gaussian avatar representations.
Results are averaged over two representative subjects under the same hardware and training settings.
}
\label{tab:transfer_cost}
\resizebox{\columnwidth}{!}{
\begin{tabular}{llccc}
\hline
Representation
& Variant
& Training Time (h)$\downarrow$
& Peak Memory (MB)$\downarrow$
& Inference Time (ms/frame)$\downarrow$ \\
\hline

\multirow{2}{*}{GaussianAvatars}
& Original
& 7.39 & 2231 & 6.12 \\

& + Fresco++
& 10.90 & 2293 & 6.00 \\
\hline

\multirow{2}{*}{TensorGA}
& Original
& 9.84 & 3363 & 6.52 \\

& + Fresco++
& 13.03 & 3479 & 6.71 \\
\hline
\end{tabular}
}
\end{table}

As shown in Table~\ref{tab:transfer_cost}, integrating Fresco++ introduces additional training time for both representations, while the increase in peak GPU memory remains small.
The additional training overhead mainly comes from the extra image-space filtering in the frequency curriculum and the multi-view feature consistency computation required by CGC.
In contrast, the inference time remains essentially unchanged, since Fresco++ does not introduce additional inference-stage modules or modify the original rendering pipeline.
These results indicate that the proposed optimization strategy improves reconstruction quality while preserving the original inference efficiency.

\section{Limitations and Future Work}

Although Fresco++ demonstrates consistent improvements across different head-avatar models, several limitations remain.
First, our current evaluation mainly focuses on head-avatar reconstruction.
Extending the proposed optimization strategy to broader dynamic human representations and more diverse scenarios would be an interesting direction for future work.

Second, Fresco++ introduces a moderate amount of additional computation during training due to frequency-aware image filtering and multi-view feature consensus.
As shown by the cost comparison in our transferability experiments, this overhead is mainly introduced during training, while the original inference pipeline and runtime efficiency are essentially preserved.
Future work could further explore more efficient training-time consistency supervision without compromising reconstruction quality.

\section{Conclusion}
In this paper, we present Fresco++, a unified optimization framework for fine-grained head avatar reconstruction.
Our approach is built upon two complementary designs.
First, we introduce a spatially localized frequency curriculum that progressively organizes optimization from low-frequency stabilization to high-frequency refinement, enabling the model to recover fine-grained appearance details in a more stable coarse-to-fine manner.
Second, we propose Canonical Group Consensus (CGC), which establishes reliable cross-view supervision through canonical structural associations and multi-view consensus, thereby improving the consistency of corresponding local regions across viewpoints.
Together, these components enhance both local detail fidelity and cross-view stability during avatar optimization.
Extensive experiments demonstrate consistent improvements over existing methods, while further transferability studies on GaussianAvatars and TensorGA show
that the proposed strategy can be effectively applied to different Gaussian
avatar representations, with additional training cost but essentially unchanged
inference efficiency.
We believe Fresco++ provides a practical and general optimization paradigm for high-quality head avatar reconstruction and offers a promising direction for more transferable and fine-grained avatar modeling.


\vfill


\begin{thebibliography}{99}

\bibitem{waggoner2009my}
Z. Waggoner, \textit{My avatar, my self: Identity in video role-playing games}. McFarland, 2009.

\bibitem{he2020collabovr}
Z. He, R. Du, and K. Perlin, ``Collabovr: A reconfigurable framework for creative collaboration in virtual reality,'' in \textit{2020 IEEE International Symposium on Mixed and Augmented Reality (ISMAR)}, pp. 542--554, 2020.

\bibitem{zhang2026fresco}
S. Zhang, Y. Li, Y. Wang, Q. Ke, and C. Chen,
``Fresco: Frequency-spatial consistent optimization for fine-grained head avatar modeling,''
in \textit{Proceedings of the IEEE/CVF Conference on Computer Vision and Pattern Recognition},
pp. 40130--40139, 2026.

\bibitem{lombardi2018deep}
S. Lombardi, J. Saragih, T. Simon, and Y. Sheikh, ``Deep appearance models for face rendering,'' \textit{ACM Transactions on Graphics (ToG)}, vol. 37, no. 4, pp. 1--13, 2018.

\bibitem{orts2016holoportation}
S. Orts-Escolano, C. Rhemann, S. Fanello, W. Chang, A. Kowdle, Y. Degtyarev, D. Kim, P. L. Davidson, S. Khamis, M. Dou, et al., ``Holoportation: Virtual 3d teleportation in real-time,'' in \textit{Proceedings of the 29th annual symposium on user interface software and technology}, pp. 741--754, 2016.

\bibitem{barron2021mip}
J. T. Barron, B. Mildenhall, M. Tancik, P. Hedman, R. Martin-Brualla, and P. P. Srinivasan, ``Mip-nerf: A multiscale representation for anti-aliasing neural radiance fields,'' in \textit{Proceedings of the IEEE/CVF international conference on computer vision}, pp. 5855--5864, 2021.

\bibitem{chen2022tensorf}
A. Chen, Z. Xu, A. Geiger, J. Yu, and H. Su, ``Tensorf: Tensorial radiance fields,'' in \textit{European conference on computer vision}, pp. 333--350, 2022.

\bibitem{mildenhall2021nerf}
B. Mildenhall, P. P. Srinivasan, M. Tancik, J. T. Barron, R. Ramamoorthi, and R. Ng, ``Nerf: Representing scenes as neural radiance fields for view synthesis,'' \textit{Communications of the ACM}, vol. 65, no. 1, pp. 99--106, 2021.

\bibitem{muller2022instant}
T. M{\"u}ller, A. Evans, C. Schied, and A. Keller, ``Instant neural graphics primitives with a multiresolution hash encoding,'' \textit{ACM transactions on graphics (TOG)}, vol. 41, no. 4, pp. 1--15, 2022.

\bibitem{hong2022headnerf}
Y. Hong, B. Peng, H. Xiao, L. Liu, and J. Zhang, ``Headnerf: A real-time nerf-based parametric head model,'' in \textit{Proceedings of the IEEE/CVF Conference on Computer Vision and Pattern Recognition}, pp. 20374--20384, 2022.

\bibitem{wang2023neural}
C. Wang, D. Kang, Y. Cao, L. Bao, Y. Shan, and S. Zhang, ``Neural point-based volumetric avatar: Surface-guided neural points for efficient and photorealistic volumetric head avatar,'' in \textit{SIGGRAPH Asia 2023 Conference Papers}, pp. 1--12, 2023.

\bibitem{xu2023avatarmav}
Y. Xu, L. Wang, X. Zhao, H. Zhang, and Y. Liu, ``Avatarmav: Fast 3d head avatar reconstruction using motion-aware neural voxels,'' in \textit{ACM SIGGRAPH 2023 Conference Proceedings}, pp. 1--10, 2023.

\bibitem{wang2021neus}
P. Wang, L. Liu, Y. Liu, C. Theobalt, T. Komura, and W. Wang, ``Neus: Learning neural implicit surfaces by volume rendering for multi-view reconstruction,'' \textit{arXiv preprint arXiv:2106.10689}, 2021.

\bibitem{yariv2021volume}
L. Yariv, J. Gu, Y. Kasten, and Y. Lipman, ``Volume rendering of neural implicit surfaces,'' \textit{Advances in neural information processing systems}, vol. 34, pp. 4805--4815, 2021.

\bibitem{dhamo2024headgas}
H. Dhamo, Y. Nie, A. Moreau, J. Song, R. Shaw, Y. Zhou, and E. P{\'e}rez-Pellitero, ``Headgas: Real-time animatable head avatars via 3d gaussian splatting,'' in \textit{European Conference on Computer Vision}, pp. 459--476, 2024.

\bibitem{qian2024gaussianavatars}
S. Qian, T. Kirschstein, L. Schoneveld, D. Davoli, S. Giebenhain, and M. Nie{\ss}ner, ``Gaussianavatars: Photorealistic head avatars with rigged 3d gaussians,'' in \textit{Proceedings of the IEEE/CVF Conference on Computer Vision and Pattern Recognition}, pp. 20299--20309, 2024.

\bibitem{wang2025gaussianhead}
J. Wang, J. Xie, X. Li, F. Xu, C. Pun, and H. Gao, ``Gaussianhead: High-fidelity head avatars with learnable gaussian derivation,'' \textit{IEEE Transactions on Visualization and Computer Graphics}, 2025.

\bibitem{xiang2024flashavatar}
J. Xiang, X. Gao, Y. Guo, and J. Zhang, ``Flashavatar: High-fidelity head avatar with efficient gaussian embedding,'' in \textit{Proceedings of the IEEE/CVF Conference on Computer Vision and Pattern Recognition}, pp. 1802--1812, 2024.

\bibitem{gafni2021dynamic}
G. Gafni, J. Thies, M. Zollhofer, and M. Nie{\ss}ner, ``Dynamic neural radiance fields for monocular 4d facial avatar reconstruction,'' in \textit{Proceedings of the IEEE/CVF Conference on Computer Vision and Pattern Recognition}, pp. 8649--8658, 2021.

\bibitem{zheng2022avatar}
Y. Zheng, V. F. Abrevaya, M. C. B{\"u}hler, X. Chen, M. J. Black, and O. Hilliges, ``Im avatar: Implicit morphable head avatars from videos,'' in \textit{Proceedings of the IEEE/CVF conference on computer vision and pattern recognition}, pp. 13545--13555, 2022.

\bibitem{li2017learning}
T. Li, T. Bolkart, M. J. Black, H. Li, and J. Romero, ``Learning a model of facial shape and expression from 4D scans,'' \textit{ACM Trans. Graph.}, vol. 36, no. 6, pp. 194--1, 2017.

\bibitem{grassal2022neural}
P. Grassal, M. Prinzler, T. Leistner, C. Rother, M. Nie{\ss}ner, and J. Thies, ``Neural head avatars from monocular rgb videos,'' in \textit{Proceedings of the IEEE/CVF conference on computer vision and pattern recognition}, pp. 18653--18664, 2022.

\bibitem{zheng2023pointavatar}
Y. Zheng, W. Yifan, G. Wetzstein, M. J. Black, and O. Hilliges, ``Pointavatar: Deformable point-based head avatars from videos,'' in \textit{Proceedings of the IEEE/CVF conference on computer vision and pattern recognition}, pp. 21057--21067, 2023.

\bibitem{guedon2024sugar}
A. Gu{\'e}don and V. Lepetit, ``Sugar: Surface-aligned gaussian splatting for efficient 3d mesh reconstruction and high-quality mesh rendering,'' in \textit{Proceedings of the IEEE/CVF Conference on Computer Vision and Pattern Recognition}, pp. 5354--5363, 2024.

\bibitem{wang2025mega}
C. Wang, D. Kang, H. Sun, S. Qian, Z. Wang, L. Bao, and S. Zhang, ``Mega: Hybrid mesh-gaussian head avatar for high-fidelity rendering and head editing,'' in \textit{Proceedings of the Computer Vision and Pattern Recognition Conference}, pp. 26274--26284, 2025.

\bibitem{xu2024gaussian}
Y. Xu, B. Chen, Z. Li, H. Zhang, L. Wang, Z. Zheng, and Y. Liu, ``Gaussian head avatar: Ultra high-fidelity head avatar via dynamic gaussians,'' in \textit{Proceedings of the IEEE/CVF conference on computer vision and pattern recognition}, pp. 1931--1941, 2024.

\bibitem{cai2024hybrid}
H. Cai, Y. Xiao, X. Wang, J. Li, Y. Guo, Y. Fan, S. Gao, and J. Zhang, ``Hybrid Explicit Representation for Ultra-Realistic Head Avatars,'' \textit{arXiv preprint arXiv:2403.11453}, 2024.

\bibitem{zhang2025eavatar}
S. Zhang, C. Chen, Y. Wang, Q. Ke, and Y. Li, ``EAvatar: Expression-Aware Head Avatar Reconstruction with Generative Geometry Priors,'' \textit{arXiv preprint arXiv:2508.13537}, 2025.

\bibitem{lin2021barf}
C. Lin, W. Ma, A. Torralba, and S. Lucey, ``Barf: Bundle-adjusting neural radiance fields,'' in \textit{Proceedings of the IEEE/CVF international conference on computer vision}, pp. 5741--5751, 2021.

\bibitem{yang2023freenerf}
J. Yang, M. Pavone, and Y. Wang, ``Freenerf: Improving few-shot neural rendering with free frequency regularization,'' in \textit{Proceedings of the IEEE/CVF conference on computer vision and pattern recognition}, pp. 8254--8263, 2023.

\bibitem{zhang2024fregs}
J. Zhang, F. Zhan, M. Xu, S. Lu, and E. Xing, ``Fregs: 3d gaussian splatting with progressive frequency regularization,'' in \textit{Proceedings of the IEEE/CVF Conference on Computer Vision and Pattern Recognition}, pp. 21424--21433, 2024.

\bibitem{thies2019deferred}
J. Thies, M. Zollh{\"o}fer, and M. Nie{\ss}ner, ``Deferred neural rendering: Image synthesis using neural textures,'' \textit{Acm Transactions on Graphics (TOG)}, vol. 38, no. 4, pp. 1--12, 2019.

\bibitem{kirschstein2023nersemble}
T. Kirschstein, S. Qian, S. Giebenhain, T. Walter, and M. Nie{\ss}ner, ``Nersemble: Multi-view radiance field reconstruction of human heads,'' \textit{ACM Transactions on Graphics (TOG)}, vol. 42, no. 4, pp. 1--14, 2023.

\bibitem{lee2020maskgan}
C. Lee, Z. Liu, L. Wu, and P. Luo, ``Maskgan: Towards diverse and interactive facial image manipulation,'' in \textit{Proceedings of the IEEE/CVF conference on computer vision and pattern recognition}, pp. 5549--5558, 2020.

\bibitem{over2021processing}
J. R. Over, A. C. Ritchie, C. J. Kranenburg, J. A. Brown, D. D. Buscombe, T. Noble, C. R. Sherwood, J. A. Warrick, and P. A. Wernette, ``Processing coastal imagery with Agisoft Metashape Professional Edition, version 1.6—Structure from motion workflow documentation,'' US Geological Survey, 2021.

\bibitem{park2021nerfies}
K. Park, U. Sinha, J. T. Barron, S. Bouaziz, D. B. Goldman, S. M. Seitz, and R. Martin-Brualla, ``Nerfies: Deformable neural radiance fields,'' in \textit{Proceedings of the IEEE/CVF international conference on computer vision}, pp. 5865--5874, 2021.

\bibitem{kerbl20233d}
B. Kerbl, G. Kopanas, T. Leimk{\"u}hler, and G. Drettakis, ``3D Gaussian splatting for real-time radiance field rendering,'' \textit{ACM Trans. Graph.}, vol. 42, no. 4, pp. 139--1, 2023.

\bibitem{liao2025hhavatar}
Z. Liao, Y. Xu, Z. Li, Q. Li, B. Zhou, R. Bai, D. Xu, H. Zhang, and Y. Liu, ``HHAvatar: Gaussian head avatar with dynamic hairs,'' \textit{IEEE Transactions on Pattern Analysis and Machine Intelligence}, 2025.

\bibitem{xu2025gphm}
Y. Xu, Z. Su, Q. Wu, and Y. Liu, ``Gphm: Gaussian parametric head model for monocular head avatar reconstruction,'' \textit{IEEE Transactions on Pattern Analysis and Machine Intelligence}, 2025.

\bibitem{wang20253d}
Y. Wang, X. Wang, R. Yi, Y. Fan, J. Hu, J. Zhu, and L. Ma, ``3D Gaussian Head Avatars with Expressive Dynamic Appearances by Compact Tensorial Representations,'' in \textit{Proceedings of the Computer Vision and Pattern Recognition Conference}, pp. 21117--21126, 2025.

\bibitem{qin2024high}
M. Qin, Y. Liu, Y. Xu, X. Zhao, Y. Liu, and H. Wang, ``High-fidelity 3d head avatars reconstruction through spatially-varying expression conditioned neural radiance field,'' in \textit{Proceedings of the AAAI Conference on Artificial Intelligence}, vol. 38, no. 5, pp. 4569--4577, 2024.

\bibitem{xie2024g}
S. Xie, S. Zhou, K. Sakurada, R. Ishikawa, M. Onishi, and T. Oishi, ``G 2 f R: Frequency Regularization in Grid-Based Feature Encoding Neural Radiance Fields,'' in \textit{European Conference on Computer Vision}, pp. 186--203, 2024.

\bibitem{li2024spectralnerf}
R. Li, J. Liu, G. Liu, S. Zhang, B. Zeng, and S. Liu, ``Spectralnerf: Physically based spectral rendering with neural radiance field,'' in \textit{Proceedings of the AAAI Conference on Artificial Intelligence}, vol. 38, no. 4, pp. 3154--3162, 2024.

\bibitem{bengio2009curriculum}
Y. Bengio, J. Louradour, R. Collobert, and J. Weston, ``Curriculum learning,'' in \textit{Proceedings of the 26th annual international conference on machine learning}, pp. 41--48, 2009.

\bibitem{lu2025lsf}
X. Lu, C. Zhuang, C. Jin, Z. Lu, Y. Wang, W. Liu, and J. Xiao, ``LSF-Animation: Label-Free Speech-Driven Facial Animation via Implicit Feature Representation,'' in \textit{Proceedings of the SIGGRAPH Asia 2025 Conference Papers}, pp. 1--12, 2025.

\bibitem{zhang2023sdrnet}
S. Zhang, F. Song, G. Song, and M. Yang, ``Sdrnet: Shape decoupled regression network for 3d face reconstruction,'' in \textit{ICASSP 2023-2023 IEEE International Conference on Acoustics, Speech and Signal Processing (ICASSP)}, pp. 1--5, 2023.

\bibitem{kabadayi2026physhead}
B. Kabadayi, V. Sklyarova, W. Zielonka, J. Thies, and G. Pons-Moll,
``PhysHead: Simulation-ready Gaussian head avatars,''
\textit{arXiv preprint arXiv:2604.06467}, 2026.

\bibitem{liu2025tetgs}
H. Liu, Y. Men, and Z. Lian,
``Creating your editable 3D photorealistic avatar with tetrahedron-constrained Gaussian splatting,''
in \textit{2025 IEEE/CVF Conference on Computer Vision and Pattern Recognition (CVPR)},
pp. 15976--15986, 2025.

\bibitem{zhang2025hravatar}
D. Zhang, Y. Liu, L. Lin, Y. Zhu, K. Chen, M. Qin, Y. Li, and H. Wang,
``HRAvatar: High-quality and relightable Gaussian head avatar,''
in \textit{2025 IEEE/CVF Conference on Computer Vision and Pattern Recognition (CVPR)},
pp. 26285--26296, 2025.

\bibitem{feng2025gpavatar}
W.-Q. Feng, D. Han, Z.-K. Zhou, S. Li, X. Liu, P. Wan, D. Zhang, and M. Wang,
``GPAvatar: High-fidelity head avatars by learning efficient Gaussian projections,''
in \textit{2025 IEEE/CVF Conference on Computer Vision and Pattern Recognition (CVPR)},
pp. 250--259, 2025.

\bibitem{sun2025svghead}
H. Sun, C. Wang, T.-X. Xu, J. Huang, D. Kang, C. Guo, and S.-H. Zhang,
``SVG-Head: Hybrid surface-volumetric Gaussians for high-fidelity head reconstruction and real-time editing,''
in \textit{2025 IEEE/CVF International Conference on Computer Vision (ICCV)},
pp. 13326--13335, 2025.

\bibitem{tang2025gaf}
J. Tang, D. Davoli, T. Kirschstein, L. Schoneveld, and M. Niessner,
``GAF: Gaussian avatar reconstruction from monocular videos via multi-view diffusion,''
in \textit{2025 IEEE/CVF Conference on Computer Vision and Pattern Recognition (CVPR)},
pp. 5546--5558, 2025.

\bibitem{ren2026omgavatar}
J. Ren, L. Liu, and S. Hoi,
``OMG-Avatar: One-shot Multi-LOD Gaussian head avatar,''
in \textit{Proceedings of the IEEE/CVF Conference on Computer Vision and Pattern Recognition (CVPR)},
pp. 11017--11028, 2026.

\bibitem{chharia2026mvchead}
A. Chharia and F. De la Torre,
``Multi-view consistent 3D Gaussian head avatars without multi-view generation,''
in \textit{Proceedings of the IEEE/CVF Conference on Computer Vision and Pattern Recognition (CVPR)},
pp. 40163--40174, 2026.

\bibitem{prinzler2026match}
M. Prinzler, P. Gotardo, S. Tang, and T. Bolkart,
``Feed-forward Gaussian registration for head avatar creation and editing,''
in \textit{Proceedings of the IEEE/CVF Conference on Computer Vision and Pattern Recognition (CVPR)},
pp. 25270--25280, 2026.

\bibitem{song2026progressiveavatars}
K. Song, J. Cui, and J. Zhang,
``ProgressiveAvatars: Progressive animatable 3D Gaussian avatars,''
in \textit{Proceedings of the IEEE/CVF Conference on Computer Vision and Pattern Recognition (CVPR)},
pp. 32518--32527, 2026.

\end{thebibliography}
\end{document}